\newif\ifllncs
\llncsfalse

\documentclass[letterpaper,11pt]{article}
\usepackage{times}
\usepackage{amsmath}

\usepackage{amsthm}
\usepackage[numbers,square,sort&compress]{natbib}
\usepackage[pdfstartview=FitH,colorlinks,linkcolor=blue,filecolor=blue,citecolor=blue,urlcolor=blue,pagebackref=true]{hyperref}

\usepackage{amsmath,amssymb,mathtools}
\usepackage{framed,color,url}
\usepackage{standalone}

\usepackage{caption}

\usepackage{fullpage}
\usepackage{algorithm2e,algorithmicx,algpseudocode,tikz,wrapfig}

\usepackage{graphicx}               
\usepackage{cite} 
\usepackage{nicefrac} 
\usepackage{paralist} 
\usepackage{boxedminipage} 
\usepackage{upgreek}

\usepackage[makeroom]{cancel}
\definecolor{RoyalBlue}{cmyk}{1, 0.50, 0, 0}
\definecolor{ForestGreen}{cmyk}{0.864, 0.0, 0.429, 0.396}
\definecolor{Brown}{cmyk}{0.0,0.692,0.925,0.529}

\usepackage{centernot}

\newcommand{\afhat}[2]{\tilde{f}_\xi [\pfix{#1}{#2}]}
\newcommand{\pkrvar}{D_{p,k,\rho}}
\newcommand{\pkrvarp}{D_{p,k,\rho'}}

\newcommand{\problem}{\ensuremath{\mathsf{P}}\xspace}
\newcommand{\pRes}{p\text{-}\mathsf{Res}}

\newcommand{\Rplus}{\RR_+}
\newcommand{\Risk}{\mathrm{Risk}}

\newcommand{\Rej}{\mathsf{R}}

\newcommand{\pTam}{p\text{-}\mathsf{Tam}}

\newcommand{\pfix}[2]{ #1_{\leq #2}}

\newcommand{\fhat}[2]{\ifthenelse{\equal{#2}{}}{\hat{f}[#1]}{\ifthenelse{\equal{#2}{0}}{\hat{f}[\emptyset]}{\hat{f}[#1_{\leq #2}]}}}
\newcommand{\gain}[2]{\ifthenelse{\equal{#2}{}}{g[#1]}{g[#1_{\leq #2}]}}

\newcommand{\pr}[2][]{\Pr_{\ifthenelse{\isempty{#1}}{}{{#1}}}\left[{#2}\right]}

\newcommand{\pmint}{[-1,+1]}

\newcommand{\Tamp}{\mathsf{Tam}}

\newcommand{\tam}{\mathrm{tam}}
\newcommand{\res}{\mathrm{res}}
\newcommand{\bud}{\mathrm{bud}}

\newcommand{\parag}[1]{\paragraph{#1.}}
\makeatother

\newcommand{\Bad}{\mathsf{Bad}}
\newcommand{\Good}{\mathsf{Good}}

\newcommand{\e}{\mathrm{e}}

\newcommand{\loss}{\mathrm{Loss}}
\newcommand{\Loss}{\loss}
\newcommand{\Adv}{\mathsf{A}}
\newcommand{\adv}{\Adv}

\newcommand{\advC}{\mathcal{A}}

\usepackage{graphicx,accents}

\newcommand{\remove}[1]{}

\newcommand{\es}{\emptyset} 
\newcommand{\se}{\subseteq}

\newcommand{\mal}[1]{\ensuremath{\widehat{#1}}\xspace} 

\newcommand{\ceil}[1]{\lceil #1 \rceil}

\newcommand{\set}[1]{\{ #1 \}}

\newcommand{\bits}{\{0,1\}}

\newcommand{\To}{\mapsto}

\newcommand{\R}{{\mathbb R}}
\newcommand{\N}{{\mathbb N}}

\newcommand{\cD}{{\mathcal D}}

\newcommand{\cS}{{\mathcal S}}

\newcommand{\cX}{{\mathcal X}}

\usepackage{bm}

\newcommand{\sfE}{\mathsf{E}}

\newcommand{\eps}{\varepsilon}

\newcommand{\poly}{\operatorname{poly}}

\newcommand{\Exp}{\operatorname*{\mathbf{E}}}
\newcommand{\Ex}{\Exp}

\newcommand{\Var}{\operatorname*{Var}}

\newcommand{\Supp}{\operatorname{Supp}}

\ifllncs
\else
\newtheorem{theorem}{Theorem}[section]
\theoremstyle{plain}

\newtheorem{proposition}[theorem]{Proposition}

\theoremstyle{definition}

\theoremstyle{definition}
\newtheorem{remark}[theorem]{Remark}

\fi

\newcommand{\sdotfill}{\textcolor[rgb]{0.8,0.8,0.8}{\dotfill}} 

\newtheorem{proto}[theorem]{Protocol}
\newtheorem{protoc}[theorem]{Protocol}

\newcommand{\namedref}[2]{#1~\ref{#2}}

\newcommand{\torestate}[3]{%
\expandafter \def \csname BBRESTATE #2 \endcsname{#3}
\theoremstyle{plain}
\newtheorem{BBRESTATETHMNUM#2}[theorem]{#1}
\begin{BBRESTATETHMNUM#2}\label{#2}\csname BBRESTATE #2 \endcsname   \end{BBRESTATETHMNUM#2}
\newtheorem*{BBRESTATETHMNONNUM#2}{\namedref{#1}{#2}}
}

\newcommand{\restate}[1]{\begin{BBRESTATETHMNONNUM#1}[Restated] \csname BBRESTATE #1 \endcsname
\end{BBRESTATETHMNONNUM#1}}

\usepackage{xspace}
\newcommand{\XX}{\ensuremath{\mathcal{X}}\xspace} 
\newcommand{\YY}{\ensuremath{\mathcal{Y}}\xspace} 

\newcommand{\DD}{\ensuremath{D}\xspace} 
\newcommand{\dist}{\DD} 
\newcommand{\distC}{\ensuremath{\mathcal{D}}\xspace}

\newcommand{\prob}[1]{\ensuremath{\Pr\left[ #1 \right]}\xspace}

\newcommand{\expected}[1]{\ensuremath{\mathbf{E}\left[#1\right]}\xspace}
\newcommand{\expectedsub}[2]{\ensuremath{\mathbf{E}_{#1}\left[#2\right]}\xspace}

\newcommand{\error}{\mathrm{Err}}

\newcommand{\hypothesisc}{\ensuremath{\mathcal H}\xspace}
\newcommand{\hypoC}{\hypothesisc}

\newcommand{\hconcept}{\ensuremath{h}\xspace}

\newcommand{\RR}{\ensuremath{\mathbb R}\xspace}

\newcounter{definitioncnt}
\newtheorem{ourdefinition}[definitioncnt]{Definition}

\newtheorem{ourconstruction}[definitioncnt]{Construction}
\newcounter{thmcnt}
\newtheorem{ourclaim}[thmcnt]{Claim}
\newtheorem{ourlemma}[thmcnt]{Lemma}
\newtheorem{ourproposition}[thmcnt]{Proposition}
\newtheorem{ourtheorem}[thmcnt]{Theorem}
\newtheorem{ourcorollary}[thmcnt]{Corollary}

\newcommand{\Mnote}[1]{{\color{purple} [\bf {Mahmoody:}  #1]}}
\newcommand{\Snote}[1]{{\color{green} [\bf {Saeed:}  #1]}}
\newcommand{\Dnote}[1]{{\color{ForestGreen} [\bf {Dimitris:}  #1]}}

\begin{document}

\title{Learning under $p$-Tampering Attacks}

\author{Saeed Mahloujifar\thanks{University of Virginia, \texttt{saeed@virginia.edu}. Supported by  University of Virginia's SEAS Research Innovation Award.} \and Dimitrios I.~Diochnos\thanks{University of Virginia, \texttt{diochnos@virginia.edu}.} \and Mohammad Mahmoody\thanks{University of Virginia, \texttt{mohammad@virginia.edu}. Supported by NSF CAREER award CCF-1350939  and University of Virginia's SEAS Research Innovation Award.}}

\date{}
\maketitle

\begin{abstract}
Recently, Mahloujifar and Mahmoody (TCC'17) studied attacks against learning algorithms  using a special case of Valiant's malicious noise, called $p$-tampering,  in which the adversary gets to change any training example with independent probability $p$ but  is limited to only choose `adversarial' examples with correct labels. They obtained $p$-tampering attacks that increase the error probability in the so called `targeted' poisoning model in which the adversary's goal is to increase the loss of the trained hypothesis over a particular test example. At the heart of their attack was an efficient algorithm to bias the expected value of any bounded real-output function through $p$-tampering. 

In this work, we present new biasing attacks for increasing the expected value of bounded real-valued functions.  Our improved biasing attacks, directly imply improved $p$-tampering attacks against learners in the targeted poisoning model. As a bonus, our attacks come with considerably simpler analysis.
We also  study the possibility of PAC learning under $p$-tampering attacks in the \emph{non-targeted} (aka indiscriminate) setting where the adversary's goal is  to increase the risk of the generated hypothesis (for a random test example). We show that PAC learning is  \emph{possible} under $p$-tampering poisoning attacks essentially whenever it is possible in the realizable setting without the attacks. We further show that PAC learning  under `no-mistake' adversarial noise is \emph{not} possible, if the adversary could choose the (still limited to only $p$ fraction of) tampered examples that  she substitutes with adversarially chosen ones. Our formal model for such `bounded-budget' tampering attackers is  inspired by the notions of (strong) adaptive corruption in secure multi-party computation.


\end{abstract}

\newpage
\tableofcontents



\newpage
\section{Introduction}
In his seminal work,  Valiant \citep{Valiant:PAC} introduced the Probably Approximately Correct (PAC) model of learning that triggered a significant amount of work in the theory of machine learning.\footnote{The original model studies learnability in a distribution-free sense, but it  also make sense for classes of distributions; 
\citep{BenedekItai::Fixed}.} 
%
An important characteristic of learning algorithms is their ability to cope with noise. 
Valiant also initiated a study of adversarial noise \citep{Valiant::DisjunctionsConjunctions}  in  which each incoming training example is chosen, with independent probability $p$, by an adversary who knows the learning algorithm. 
Since no assumptions are made on such adversarial examples, this type of noise is called \emph{malicious}.
Subsequently, Kearns and Li \citep{KearnsLi::Malicious} and the follow-up work of Bshouty et. al \citep{NastyNoise} 
essentially proved impossibility of PAC learning under such 
malicious noise by heavily relying on the existence of \emph{mistakes} (i.e.,~wrong labels) 
in adversarial examples given to the learner under a carefully chosen specific distribution.
In its simplest form, the main idea of their approach was to make it impossible for the learner to distinguish between two different target concepts,
and this was achieved by 
generating wrong labels at an appropriate rate under a carefully chosen pathological distribution.
This approach for obtaining a negative result 
is a consequence of Valiant's model of distribution-free PAC learning,
since in general, the learning algorithms have to be able to deal with  well with all distributions.\footnote{
    In fact, determining properties of \emph{distribution-free} learning algorithms by looking at their behavior 
    \emph{under specific distributions} makes sense in the noise-free setting as well;
    for example, \citep{FourGermans,PAC:LowerBound} obtain lower bounds on the number 
    of examples needed for learning by looking at specific distributions.
} 

The method of induced distributions gained popularity and was seen as a tool 
that was 
used in order to prove 
negative results within various noise models.
Sloan in \citep{Sloan::Noise:four-types} used this method 
in order to determine an upper bound on the error rate that can be tolerated in 
a noise model where the labels can be mislabeled maliciously.
Bshouty, Eiron, and Kushilevitz
\citep{NastyNoise}
studied a noise model closely related to Valiant's malicious noise,  
in which  the adversary is allowed to make its choices based on the full knowledge of the original training examples;
in their work they used the method of induced distributions 
in order to give an upper bound on the maximum amount of noise that can be tolerated 
by any learning algorithm. 

In contrast to the works of 
\citep{KearnsLi::Malicious,Sloan::Noise:four-types,NastyNoise} 
who used the method of (pathological) induced distributions 
from where the malicious samples were drawn, 
in this work we are interested in attackers who do 
\emph{not} have any control over the the original distributions, 
but they can choose and inject malicious examples in certain (restricted) ways. 
On the other hand, 
it is also worth noting that near the end of our work in this paper 
we also provide a construction for a negative result within PAC learning.
Interestingly, our idea for the behavior of the adversary that 
yields this negative learning result in our framework, 
is the same as the key idea underlying the method of induced distributions  
where one tries to make it impossible for the learner 
to disambiguate between competing target concepts;
however, in our context no wrong labels are used.

%

\parag{Poisoning attacks} Impossibility results against learning under adversarial noise  could be seen as attacks against learners in which the attacker injects some malicious training examples to the training set and tries to prevent the learner from finding a hypothesis with low risk. Such attackers, in general, are studied in the context of \emph{poisoning}  (a.k.a causative) attacks\footnote{At a technical level, the malicious noise model also allows the adversary to know the \emph{full} state (and thus the randomness) of the learner, while this knowledge is not given to the adversary of the poisoning attacks, who might be limited in  other ways as well.} \citep{barreno2006can,biggio2012poisoning,papernot2016towards}. Such attacks could happen naturally when a learning process happens over time \citep{rubinstein2009stealthy,rubinstein2009antidote} 
and the adversary has some noticeable chance of injecting or substituting malicious training data in an online manner. A stronger form of poisoning attacks are the so called \emph{targeted} (poisoning) attacks \citep{barreno2006can,shen2016uror}, where the adversary performs the poisoning attack while she has a particular test example in mind, and her  goal is to make the final generated hypothesis fail on that particular test example. While poisoning attacks  against \emph{specific} learners were studied before \citep{awasthi2014power,xiao2015feature,shen2016uror}, the recent work of Mahloujifar and Mahmoody \citep{pTampTCC17} presented a generic  targeted poisoning attack that could adapt to apply to  \emph{any learner},  so long as there is an initial non-negligible  error over the target point.

\parag{$p$-tampering attacks} The work of  \citep{pTampTCC17} proved their result using a special case  of Valiant's malicious noise, called $p$-tampering, in which the attacker can only use \emph{mistake-free malicious noise}. Namely, similar to Valiant's model, any incoming training example might be chosen adversarially with independent probability $p$ (see Definition \ref{def:pTamp} for a formalization). However, the difference between $p$-tampering noise and Valiant's malicious noise  (and even from all of its special cases studied before \citep{Sloan::Noise:four-types}) is that a $p$-tampering adversary is only allowed to choose \emph{valid} tampered examples with \emph{correct} labels\footnote{This is assuming that the original training distribution only contains correct labels.} to substitute the original examples. 
As such, although the attributes can change pretty much arbitrarily in the tampered examples, the label of the tampered examples shall still reflect the correct label. For example, 
the adversary can repeatedly present the same example to the learner, thus reducing the effective sample size, or it can be the case that 
the adversary 
returns correct examples that are somehow chosen against the learner's algorithm and
based on the whole history of the examples so far.
Therefore, as opposed to the general model of Valiant's malicious noise, $p$-tampering noise/attacks are `defensible' as the adversary can always claim that a malicious training example is indeed generated from the same original distribution from which the rest of the training examples are generated. Similar notions of defensible attacks are previously explored in the context of cryptographic attacks \citep{EPRINT:HIKLP10,aumann2007security}. 
Therefore, learning under $p$-tampering can be seen as a generalization of  ``robustness'' \citep{Robustness::COLT2010,DistributionChange:A,DistributionChange:B} in which the training distribution can \emph{adaptively} and \emph{adversarially} deviate from the testing distribution without using wrong labels.



\parag{Biasing bounded functions}  At the heart of the poisoning attacks  of \citep{pTampTCC17} against learners was a  $p$-tampering attack for the more  basic task of \emph{biasing} the expected value  of bounded real-valued functions. In particular, \citep{pTampTCC17} proved that for any (polynomial time computable) function $f$ mapping inputs drawn from distributions like $S \equiv D^n$ (consisting of $n$ iid `blocks') to  $[0,1]$, there is always a \emph{polynomial time}  $p$-tampering attacker $\Adv$ who changes the input distribution $S$ into $\mal{S}$ while increasing the expected value by  at least $\frac{p}{3+5p} \cdot  \Var[f(S)]$
where $\Var[\cdot]$ is the variance.\footnote{In the original version a slightly stronger bound of  $\frac{2p}{3+4p} \cdot  \Var[f(S)]$ was claimed, though the full version \citep{cryptoeprint:2017:950} corrected this to the weaker bound $\frac{p}{3+5p}\cdot  \Var[f(S)]$} (Note that the bias shall somehow depend on $\Var[f(S)]$ since constant functions cannot be biased by changing their inputs.)   On the other hand, the work of \citep{pTampTCC17} shows that for some functions even \emph{computationally unbounded} $p$-tampering attackers (who can run in exponential time) 
cannot achieve better than $\frac{\ln(1/\mu)}{1-\mu}\cdot p\cdot \Var[f(S)]$ for all $p,\mu \in (0,1)$, if $\mu = \Ex[f(S)]$,
which because of $\lim_{\mu \to 1} \frac{\ln(1/\mu)}{1-\mu} = 1 $, it means the best possible universal constant $c$ to achieve bias $c \cdot p \cdot \Var[f(S)]$ through $p$-tampering is at most $c\leq 1$.
For the special case of \emph{Boolean} function $f(\cdot)$, or alternatively when the $p$-tampering attacker is allowed to run in \emph{exponential time}, \citep{pTampTCC17} achieved almost optimal bias of $\frac{p}{1+p\cdot \mu - p}\cdot \Var[f(S)] > p \cdot \Var[f(S)]$. Using their biasing attacks, \citep{pTampTCC17} directly obtained $p$-tampering targeted poisoning attacks with related bounds.
Therefore, a main question that remained open after \citep{pTampTCC17} and is a subject of our study is the following.
What is the maximum possible bias of real-valued functions through $p$-tampering attacks? 
Resolving this question, directly leads to improved $p$-tampering poisoning attacks against learners, 
when the loss function is real-valued.

\subsection{Our Results}
\parag{Improved $p$-tampering biasing attacks} Our main technical result in this work is to improve the efficient (polynomial-time) $p$-tampering biasing attack of \citep{pTampTCC17} to achieve the bias of
$\frac{p}{1+p\cdot \mu-p}\cdot \Var[f(S)] \geq p \cdot \Var[f(S)]$ (where $\mu = \Ex[f(S)]$ for $S\equiv D^n$ and $\Var[\cdot]$ is the variance) 
in \emph{polynomial time} and  for \emph{real-valued} bounded functions with output in $[0,1]$ (see Theorem~\ref{thm:main-form}). This main result immediately allows us to get improved polynomial-time targeted $p$-tampering attacks against learners for scenarios where the loss function is not Boolean (see Corollary \ref{cor:Targ}). As in \citep{pTampTCC17}, our attacks  apply to any learning problem $\problem$ and any learner $L$ for $\problem$ as long as $L$ has a non-zero 
initial error over a specific test example $d$.

\parag{Special case of $p$-resetting attacks} The biasing attack of \citep{pTampTCC17} has an extra property that for each input block (or training example) $d_i$, if the adversary gets to tamper with $d_i$, it either does not change $d_i$ at all, or it simply `resets' it by resampling it from the original (training) distribution $D$. In this work, we refer to such limited forms of $p$-tampering attacks as \emph{$p$-resetting} attacks. Interestingly, $p$-resetting attacks were previously studied in the work of Bentov, Gabizon, and Zuckerman~\citep{bentov2016bitcoin} in the context of (ruling out) extracting uniform randomness from Bitcoin's blockchain \citep{nakamoto2008bitcoin} when the adversary controls $p$ fraction of the computing power.\footnote{To compare the terminologies, the work of \citep{bentov2016bitcoin} studies \emph{$p$-resettable} sources of randomness, while here we study $p$-resetting attackers that generate such sources.} Bentov, et al. \citep{bentov2016bitcoin}  showed how to achieve bias $p/12$ when the original (untampered) distribution $D$ is uniform and the function $f$ is Boolean and balanced.\footnote{The running time of the $p$-resetting attacker of \citep{bentov2016bitcoin} was $\poly(n,2^{|D|})$ where $|D|$ is the length of the binary representation of any $d \gets D$. In contrast, our $p$-resetting attacks run in time $\poly(n,|D|)$.}    
As a special case of $p$-tampering attacks, $p$-resetting attacks have interesting properties that are not present in general $p$-tampering attacks. For example, if an attacker chooses its adversarial examples from a large pool by ``skipping'' some of them, then $p$-resetting attacks need a pool of about $\approx(1+p)\cdot n$, while $p$-tampering attackers might need much more. That is because, for each tampered example, the adversary simply needs to choose one out of two original correctly labeled examples, while a $p$-tampering attacker might need more samples. 
Motivated by special applications of $p$ resetting attacks and the special properties of $p$-resetting attacks,  in this work we  also study such attacks over arbitrary block distributions $D$ and  achieve bias of at least $\frac{p}{1+p\cdot \mu}\cdot \Var[f(S)]$, improving the bias of  $\frac{p}{3+5p}\cdot \Var[f(S)]$ proved in \citep{pTampTCC17}.  

\parag{PAC learning under $p$-tampering} We also study the power of $p$-tampering (and $p$-resetting) attacks in the \emph{non-targeted} setting where the adversary's goal is simply to increase the risk of the generated hypothesis.\footnote{In the targeted setting, the $\eps$ parameter of $(\eps,\delta)$-PAC learning goes away, due to the pre-selection of the target test.} In this setting, it is indeed meaningful to study the possibility (or impossibility) of PAC learning, as the test example is chosen at random. We show that in this model, $p$-tampering attacks cannot prevent PAC learnability for `realizable' settings; that is when there is always a hypothesis consistent with the training data (see Theorem \ref{thm:PossPAC-weak}). 
We further go beyond $p$-tampering attacks and study PAC learning under more powerful adversaries who might \emph{choose} the location of training examples that are tampered with but are still limited to choose   $\leq p\cdot n$ such examples. We show that PAC learning under such adversaries depends on whether the adversary makes its tampering choices \emph{before} or \emph{after} getting to see the original  sample  $d_i$. We call these two class of attacks, respectively, weak and strong $p$-budget tampering attacks (see Definition~\ref{def:pBud}).  Our notion of strong $p$-budget tampering is inspired by  notions of adaptive corruption \citep{STOC:CFGN96} and particularly \emph{strong} adaptive corruption \citep{goldwasser2015} studied in cryptographic contexts. 
Our impossibility result of PAC learnability under strong $p$-budget attacks (see Theorem \ref{thm:NoPAC-strong}) shows that PAC learning under `mistake-free' adversarial noise is \emph{not} always possible.

Finally, we would like to point out that our positive result about PAC learnability under $p$-tampering attacks (see Theorem \ref{thm:PossPAC-weak}) shows a stark contrast between the `mistake-free' adversarial noise and general malicious noise for $p >1/2$. Indeed, when the adversary can tamper with $p\approx 1/2$ fraction of the training data in an arbitrary way for a binary classification problem, it can  make the training data completely useless by always picking the labels at random from $\bits$. Such adversary will end up changing only $p\approx 1/2$ of the examples, but will make the labels independent of the features. However, as we prove in Theorem \ref{thm:PossPAC-weak}, PAC learning is possible under $p$-tampering for any constant $p<1$.


\parag{Applications beyond attacking learners} Similar to how \citep{pTampTCC17} used their biasing attacks in applications other than attacking learners, our new biasing attacks can also be used to obtain improved polynomial-time attacks for biasing the output bit of any  seedless randomness extractors \citep{von195113,ChorGo85,SanthaV86}, as well as blockwise $p$-tampering (and $p$-resetting) attacks against security of indistinguishability-based cryptographic primitives (e.g., encryption, secure computation, etc.). 
As in \citep{pTampTCC17}, our new improved biasing attacks apply to any \emph{joint} distribution (e.g., martingales) when the tampered values affect the random process in an online way. In this work, however, we focus on the case of product distributions as they suffice for getting our attacks against learners and include all the main ideas even for the general case of random processes.
 We refer the reader to the work of \citep{pTampTCC17} for the extra applications.

\parag{Recent positive results achieving algorithmic robustness} On the positive (algorithmic) side, the seminal  works of Diakonikolas et al.~\citep{diakonikolas2016robust} and Lai et al.~\citep{lai2016agnostic} showed the surprising power of algorithmic robust inference over poisoned data  with error that does not depend on the dimension of the distribution (but still depends on the fraction of poisoned data). These works led to an active line of work (e.g., see \citep{charikar2017learning,diakonikolas2017statistical,diakonikolas2018list,diakonikolas2018sever,prasad2018robust,diakonikolas2018efficient} and references therein) exploring the possibility of robust statistics over poisoned data with algorithmic guarantees. The works of \citep{charikar2017learning,diakonikolas2018list} performed \emph{list-decodable} learning, and \citep{diakonikolas2018sever,prasad2018robust} studied supervised learning. In our attacks, however, similarly to virtually all attacks in the literature (over specific learners and models) we demonstrate inherent power of poisoning attacks (that apply to \emph{any} learner and hypothesis class) to \emph{amplify} the error of classifiers starting from small and perhaps acceptable error rates, while after the attack the error probability is essentially  one. Namely, our results show that in order to resist poisoning attacks, the same algorithms should do much better in the no-attack setting, as otherwise a poisoning attacker can increase the targeted error probability significantly.
 
\subsubsection{Ideas behind our new biasing attacks and our approach}

Our new biasing attacks built upon ideas developed in previous work \citep{ReingoldVW04,DodisOnPrSa04,BeigiEG17,dodis2015privacy,bentov2016bitcoin} in the context of attacking deterministic randomness extractors from the so called Santha-Vazirani sources  \citep{SanthaV86}. In \citep{pTampTCC17} the authors generalized the idea of `half-space' sources (introduced in \citep{ReingoldVW04,DodisOnPrSa04}) to real-valued functions, using which  they showed how to find $p$-tampering biasing attacks with  bias  $\frac{p}{1+p\cdot \mu-p}\cdot \Var[f(S)]$. However, their attacks need \emph{inefficient} (i.e., super polynomial time) tampering algorithms. 
In particular, \citep{pTampTCC17} directly defined a perturbed joint distribution $\mal{S}=(\mal{D}_1,\dots,\mal{D}_n)$ of the original product distribution $S \equiv D$ such that has two properties hold: (1) $\Ex[f(\mal{S})]$ achieves the desired bias, and (2) $\Pr[\mal{S}=z] \leq c \cdot \Pr[S=z]$ for all points $z$ and sufficiently small constant $c$, meaning that $\mal{S}$ does not increase the point-wise probabilities ``too much''. It was shown in \citep{pTampTCC17} that the second property guarantees that the distribution $\mal{S}$ can be obtained from $S$ by \emph{some} tampering algorithm, but their proof was existential, namely it said nothing about the computational complexity of such tampering algorithm. Achieving the same bias \emph{efficiently}  for \emph{real-valued} functions is the main technical challenge in this work. 
\paragraph{Our approach.} At a very high level, we show how to achieve in \emph{polynomial-time} the same bias achieved in \citep{pTampTCC17} through the following two steps.
\begin{enumerate}
    \item We first show how to obtain the same exact \emph{final} distribution achieved in \citep{pTampTCC17} through \emph{local} $p$-tampering decisions that could be implemented in polynomial time using an idealized oracle $\hat{f}[\cdot]$ that provides certain information about function $f(\cdot)$. 
    \item We then, show that the idealized oracle $\hat{f}[\cdot]$ can be approximated in polynomial time, and more importantly, the $p$-tampering attack of the previous step (using idealized oracle $\hat{f}[\cdot]$) is robust to this approximation and still achieves almost the same bias.
\end{enumerate}
\paragraph{Idealized oracle $\hat{f}[\cdot]$.} Let $d_{\leq i}=(d_1,\dots,d_{i})$ be the first $i$ blocks given as input to a function $f$.\footnote{Alternatively the first $i$ training examples, when we attack learners. However,  some of  the blocks in $(d_1,\dots,d_{i})$ might be the result of previous tampering decisions.} Now, suppose the adversary gets the chance to determine the next block $d_{i+1}$ based on its knowledge of the previously generated blocks $(d_1,\dots,d_{i})$. We achieve the goal of the first step depicted above, with the help of the following oracle provided for free to the $p$-tampering attacker.



$$\fhat{d}{{ i}} = \Ex_{d_{i+1},\dots,d_n \gets D^{n-i}}[f(d_1,\dots,d_n)].$$

In other words, $\fhat{d}{{ i}}$ computes the expected value of $f$ when each of the blocks (examples) $d_{i+1},\dots,d_n$ is drawn iid from $D$, while the first $i$ blocks $d_1, \ldots, d_i$ are fixed as dictated by $d_{\leq i}$.


Although the partial averages $\fhat{d}{{ i}}$ are not \emph{exactly} computable in polynomial time, 
they can indeed be efficiently approximated within arbitrary small additive error. As we show, our attacks are also robust to such approximations, and using the approximations of $\fhat{d}{{ i}}$ (rather than their exact values) we can still bound the bias. See Sections \ref{sec:AttProof} and Section
\ref{sec:efficiency} for the details. 

\paragraph{The case of $p$-resetting attacks.} When it comes to $p$-resetting attacks, we cannot achieve the same bias that we do achieve through general $p$-tampering attacks. However, we still use the same recipe as described above. Namely, we use the idealized oracle $\fhat{d}{{ i}}$  to make careful local sampling to keep or reset a given block $d_i$, so that the final distribution has the desired bias. We then approximate the idealized oracle while arguing that the analysis is robust to this change.




\paragraph{Comparison with the polynomial-time attacker of \citep{pTampTCC17}.} As mentioned before, the work of \citep{pTampTCC17} also provides polynomial $p$-tampering attacks with weaker bounds. At a high level, the attacks  of \citep{pTampTCC17} were  simple to describe (without using the idealized oracle $\hat{f}$), while their analyses were extremely complicated and used the function $\hat{f}$ as well as a carefully chosen potential functions based on ideas from \citep{Austrin14} in which authors presented a $p$-tampering biasing attack for the special case of  uniform Boolean blocks (i.e., $D \equiv U_1$). Our new (polynomial time) attacks takes a dual approach: the analysis of our attacks are conceptually simpler, as they directly achieve the desired bias, but the description of our attacks are more complicated as they also depend on the idealized oracle $\hat{f}$.
\section{Preliminaries}\label{sec:prelims}
\parag{Notation} We  use calligraphic letters (e.g., $\cD$) for sets and capital non-calligraphic letters (e.g., $D$) for distributions. 
By $d \gets D$ we denote that $d$ is sampled from $D$. For a randomized algorithm $L(\cdot)$, by $y \gets L(x)$ we denote the randomized execution of $L$ on input $x$ outputting $y$. For joint distributions $(X,Y)$, by $(X \mid y)$ we denote the conditional distribution $(X \mid Y = y)$. By $\Supp(D) = \set{d \mid \Pr[D=d]>0}$ we denote the support set of $D$. By $D \in \cS$ we denote that $D$ always outputs in $\cS$, namely $\Supp(D) \subseteq \cS$. By $T^D(\cdot)$ we denote an algorithm $T(\cdot)$ with oracle access to a sampler for $D$. By $D \equiv G$ we denote that distributions $D,G$ are identically distributed.  By $D^n$ we denote  $n$  iid samples from $D$.
By $\eps(n) \leq \frac{1}{\poly(n)}$ we mean $\eps(n) \leq \frac{1}{n^{\Omega(1)}}$ and by $t(n) \leq \poly(n)$ we mean $t(n) \leq n^{O(1)}$.

A learning problem $\problem=(\XX,\YY,\cD,\hypoC,\loss)$ is specified by the following components. 
The set $\XX$ is the set  of possible \emph{instances}, 
\YY is the set of possible \emph{labels}, 
$\distC$ is a class of distributions containing some joint distributions $\dist \in \distC$ over $\XX \times \YY$.\footnote{By using joint distributions over $\XX \times \YY$, we jointly model a set of distributions over $\XX$ and a concept class mapping $\XX$ to $\YY$ (perhaps with noise and uncertainty).}
The set $\hypothesisc \subseteq \YY^\XX$ is
called the \emph{hypothesis space} or \emph{hypothesis class}.
We  consider \emph{loss functions}
$\loss \colon \YY \times \YY \To \Rplus$ where $\loss(y',y)$ measures how different the `prediction' $y'$ (of some possible hypothesis $h(x)=y'$) is from the true outcome $y$.\footnote{Natural loss functions such as the 0-1 loss or the square loss assign the same amount of loss for same labels computed by $h$ and $c$ regardless of $x$.}
We call a loss function \emph{bounded} if it always takes values  in $[0,1]$. 
A natural loss function for classification tasks is to use $\loss(y',y)=0$ if $y=y'$ and $\loss(y',y)=1$ otherwise.
For a given distribution $\dist \in \distC$, 
the \emph{risk} of a hypothesis $h \in \hypoC$ is the expected loss of $h$ with respect to $\dist$, namely $\Risk_\dist(h) = \expectedsub{(x,y)\gets \dist}{\loss(h(x),y)}$. 

An \emph{example} $s$ is a pair $s=(x,y)$ where $x \in \XX$ and $y \in \YY$. An example is usually sampled from a distribution $\dist$.  
A \emph{sample} set (or sequence) $\cS$ of size $n$ is a set (or sequence) of $n$ examples.
A hypothesis \hconcept is \emph{consistent} with a sample set (or sequence) $\cS$ 
if and only if $\hconcept(x) = y$ for all $(x, y)\in \cS$. 
We assume that instances, labels, and hypotheses 
are encoded as strings over some 
alphabet
such that given a hypothesis $h$ and an instance $x$,  $h(x)$ is computable in polynomial time.
\begin{ourdefinition}[Realizability]
We say that the problem $\problem=(\XX,\YY,\distC,\hypoC,\loss)$ is realizable, if for all $\dist \in \distC$, there exists an  $h \in\hypoC$ such that $\Risk_\dist(h) =  0$. 
\end{ourdefinition}
We can now define \emph{Probably Approximately Correct (PAC)} learning. Our definition is with respect to a given set of distributions $\distC$, and it can be instantiated with one distribution $\set{D} =\distC$ to get the distribution-specific case. We can also recover  the distribution-independent  scenario,  whenever the projection of $\distC$ over $\cX$ covers all distributions.
\begin{ourdefinition}[PAC Learning]\label{def:pac}
A realizable 
problem $\problem=(\XX,\YY,\distC,\hypoC,\loss)$ is $(\eps,\delta)$-PAC learnable  if there is a (possibly randomized) learning algorithm $L$ such that for every $n$ and every  $\dist\in\distC$, it holds that, 
$$\Pr_{\substack{\cS \gets D^n, h \gets L(\cS)}}[\Risk_\dist( h) \leq \eps(n)] \geq 1-\delta(n).$$
We call  $\problem$ simply PAC learnable if  $\eps(n),\delta(n) \leq 1/\poly(n)$, 
and we call it \emph{efficiently} PAC learnable if, in addition, $L$ is running in polynomial time.
\end{ourdefinition}
\begin{ourdefinition}[Average Error of a Test]\label{def:SPpac} For a problem $\problem=(\XX,\YY,\distC,\hypoC,\loss)$, a (possibly randomized) learning algorithm $L$, a fixed test sample $(x,y)=d \gets D$ for some distribution $S$ over $\Supp(D)^n$ (e.g., $S \equiv D^n$) for some $n \in \N$, the \emph{average error}\footnote{The work \citep{pTampTCC17} called the same notion the `cost' of $d$.}  of the test example $d$ (with respect to  $S,L$) is defined as,
$$\error_{S,L}(d)=\Ex_{\substack{\cS \gets S, h \gets L(\cS)}}[\loss(h(x), y)].$$
We call $\error_{S,L}= \Ex_{d \gets D} \error_{S,L}(d)$ simply the average error.
When $L$ is clear from the context, we simply write $\error_{S}(d)$ (resp. $\error_S$) to denote $\error_{S,L}(d)$ (resp. $\error_{S,L}$).
\end{ourdefinition}
It is easy to see that a realizable problem $\problem= (\XX,\YY,\distC, \hypoC, \loss)$ with bounded loss function $\loss$  is PAC  learnable if and only if there is a learner $L$ (for $\problem$) such that  its average  error $\error_S$ is bounded by a fixed $1/\poly(n)$ function for all $D \in \distC$.\footnote{Suppose $\loss(\cdot)$ is bounded (i.e., always in $[0,1]$). If $\problem$ is $(\eps,\delta)$-PAC learnable, then by a union bound, $\error_S \leq \eps+\delta$. Moreover, if $L$ is \emph{not} $(\eps,\delta)$-PAC learnable, then its average error is at least $\eps \cdot \delta$. This  means that if $L$ has average error $\gamma=\error_S$, then  $L$ is an $(\sqrt{\gamma},\sqrt{\gamma})$-PAC learner as well. 
}
\parag{Poisoning attacks} PAC learning under adversarial noise is already defined in the literature, however, poisoning attacks include  broader classes of attacks. For example, a poisoning adversary might \emph{add} adversarial examples to the training data (thus, increasing its size) or \emph{remove} some of it adversarially. 
A more powerful form of poisoning attack is the so called \emph{targeted} poisoning attack where the adversary gets to know the target test example before poisoning the training examples.
More formally, suppose $\cS = (d_1,\dots,d_n)$ is the training examples iid sampled from $D \in \distC$. 
For a poisoning attacker $\Adv$, by $\mal{\cS} \gets \Adv(\cS)$ we denote the process through which $\Adv$ generates an adversarial training set $\mal{\cS}$ based on $\cS$. Note that, this notation does not specify the exact limitations of how $\Adv$ is allowed to tamper with $\cS$, and that is part of the definition of $\Adv$. In the targeted case, the adversary $\Adv$ is also given a test example $(x,y)=d \gets \dist$. So, we would denote this by writing $\mal{\cS} \gets \Adv(d,\cS)$ to emphasize that $d$ is the test example given as input to $\Adv$. We  use calligraphic $\advC$ to denote a  \emph{class} of attacks. Note that a particular adversary $\Adv \in \advC$ might try to poison a training set $\cS$ \emph{based} on the knowledge of a problem $\problem=(\XX,\YY,\distC,\hypoC,\loss)$. On the other hand, because sometimes we would like to limit the adversary's power based on the specific distribution $D$ (e.g., by always choosing tampered data to be in $\Supp(D)$), by $\advC_{D}\se \advC$ we denote the adversary class  for a particular distribution $D$.

\begin{ourdefinition}[Learning under  poisoning] Suppose $\problem=(\XX,\YY,\distC,\hypoC,\loss)$ is a  problem, $\advC=\cup_{D \in \cD} \advC_D$ is an  adversary class, and $L$ is a (possibly randomized) learning algorithm for $\problem$.
\begin{itemize}
    \item {\bf PAC learning under poisoning.} For 
    problem $\problem$, $L$ is an $(\eps,\delta)$-PAC learning algorithm for $\problem$ under poisoning attacks of $\advC$, if for every $D \in \distC, n\in \N$, and every adversary $\adv \in \advC_D$,
    $$\Pr_{\substack{\cS \gets D^n, \textcolor{red}{\mal{\cS} \gets \Adv(\cS)}, h \gets L(\mal{\cS})}}[\Risk_\dist( h) \leq \eps(n)] \geq 1-\delta(n).$$
PAC learnability and efficient PAC learnability are then  defined similarly to Definition \ref{def:pac}.
    \item {\bf Average error under targeted poisoning.} If $\advC$ contains \emph{targeted} poisoning attackers, for a  distribution  $D \in \distC$ and an attack $\Adv \in \advC_D$, the \emph{average error} $\error^\Adv_{D^n}(d)$ for a test example $d=(x,y)$ under poisoning attacker $\adv$ is equal to $\error_{\mal{S}}(d)$ where $\mal{S} \equiv \Adv(\textcolor{red}{d},S)$ for $S \equiv D^n$.
\end{itemize}
\end{ourdefinition}
\parag{$p$-tampering attacks} We now define the specific class of poisoning attacks studied in this work. Informally speaking, $p$-tampering attacks model attackers who will manipulate the training sequence $\cS=(d_1,\dots,d_n)$ in an \emph{online}  way, meaning while tampering with $d_i$, they do not rely on the knowledge of $d_j, j>i$. Moreover, such attacks get to tamper with $d_i$ only with independent probability $p$, modeling scenarios where the tampering even is random and outside the adversary's choice. A crucial point about $p$-tampering attacks is that they always stay in $\Supp(D)$.
The formal definition follows.
\begin{ourdefinition}[$p$-tampering/resetting attacks] \label{def:pTamp} The class of $p$-tampering attacks $\advC^p_{\tam} = \cup_{D \in \cD} \advC_D$ is defined as follows. For a distribution $D \in \distC$, any $\adv \in \advC_D$ has a  (potentially randomized) tampering algorithm $\Tamp$ such that (1) given oracle access to $D$, $\Tamp^D(\cdot) \in \Supp(D)$, and (2) given any training sequence $\cS=(d_1,\dots,d_n)$,  the tampered  $\mal{\cS}=(\mal{d}_1,\dots,\mal{d}_n)$ is generated by $\adv$ inductively (over $i \in [n]$) as follows.
\begin{itemize}
\item With probability $1-p$, let $\mal{d_i} = d_i$.
\item Otherwise, (this happens with probability $p$), get 
$\mal{d_i} \gets \Tamp^D(1^n,\mal{d}_1,\dots,\mal{d}_{i-1},d_i).$
\end{itemize}
The class of \emph{$p$-resetting} attacks $\advC^p_{\res} \subset \advC^p_{\tam}$ include special cases of $p$-tampering attacks where the tampering algorithm $\Tamp$ is restricted  as follows. Either $\Tamp(1^n,\mal{d}_1,\dots,\mal{d}_{i-1},d_i)$ outputs $d_i$, or otherwise, it will 
output a \emph{fresh} sample  $d'_i \gets D$.
In the \emph{targeted} case, the adversary $\Adv_D$ and its tampering algorithm $\Tamp$ are also given the final test example $d_0 \gets D$ as extra input (that they can read but not tamper with).
An attacker $\adv_D$ is called \emph{efficient}, if its oracle-aided tampering algorithm $\Tamp^D$ runs in polynomial time.
\end{ourdefinition}
\paragraph{Subtle aspects of the definition.} Even though one can imagine a more general definition for tampering algorithms, in all the attacks of  \citep{pTampTCC17} and the attacks of this work, the tampering algorithms do \emph{not} need to know the original un-tampered values $d_1,\dots,d_{i-1}$. Since our goal here is to design $p$-tampering attacks, we use the simplified definition above, while all of our positive results still hold for the stronger version in which the tampering algorithm is given the full history of the tampering algorithm. 
Another subtle issue is about whether $d_i$ is needed to be given to the tampering algorithm.  As already noted in \citep{pTampTCC17}, when we care about $p$-tampering distributions of $D^n$,  $d_i$ is not necessary to be given to the tampering algorithm $\Tamp$, as $\Tamp$ can itself sample a copy from $D$ and treat it like $d_i$. Therefore the `stronger' form of such attacks (where $d_i$ is given) is equivalent to the `weaker' form where $d_i$ is not given. 
In fact, if $D$ is  samplable in polynomial time,  
then this equivalence holds with respect to efficient adversaries (with efficient $\Tamp$ algorithm) as well. 
In this work, for  both $p$-tampering and $p$-resetting attacks we choose to always give $d_i$ to $\Tamp$.
Interestingly, as we will see in Section \ref{sec:PAC}, if the adversary can \emph{choose} the $~p \cdot n$ locations of tampering, the weak and strong attackers will have different powers!

\subsection{Concentration Bounds}

\begin{ourdefinition}[Hoeffding inequality \citep{Hoeffding}]\label{lem:hoeffding}
Let $X_1, \ldots, X_n$ be $n$ independent random variables where  $\Supp(X_i) \se [0,1]$ for all $i \in [n]$.
Let $X = \frac{1}{n}\sum_{i=1}^n X_i$ and 
$\lambda = \expected{X}$. Then, for any
$\xi \geq 0$,
$$\prob{\left|X - \lambda\right| \geq \xi} \leq 2\e^{-2n\xi^2} \; .$$
\end{ourdefinition}

\remove{
\Mnote{Do we use the 2nd inequality of Chernoff too?}\Snote{I think we don't use it.}
}
\begin{ourdefinition}[Chernoff Bound \citep{Chernoff}]\label{lem:chernoff}
Let $X_1, \ldots, X_n$ be $n$ independent boolean  random variables,  $\Supp(X_i) \se \bits$ for all $i \in [n]$.
Let $X = \frac{1}{n}\sum_{i=1}^n X_i$ and $\lambda = \expected{X}$. Then, for any
$\gamma \in [0,1]$,
$$\prob{X \geq (1+\gamma)\cdot \lambda} \leq \e^{-n \cdot \lambda\cdot \gamma^2/3} \; ,$$
\remove{$$\prob{X \leq (1-\gamma)\cdot \lambda} \leq \e^{-n \cdot \lambda\cdot  \gamma^2/2} \; .$$}
\end{ourdefinition}



\section{Improved \ensuremath{p}-Tampering and \ensuremath{p}-Resetting Poisoning Attacks} \label{sec:limits}
In this section we study the power of $p$-tampering attacks in the targeted setting and improve upon the $p$-tampering and $p$-resetting attacks of \citep{pTampTCC17}.  Our main tool is the following theorem giving new improved $p$-tampering and $p$-resetting attacks to bias the output of bounded real-valued functions.

\subsection{The Statement of Results}

\begin{ourtheorem}[Improved biasing attacks] \label{thm:main-form}
Let $D$ be any distribution, $S \equiv D^n$, and $f \colon \Supp(S) \to [0,1]$. Suppose $\mu = \Ex[f(S)]$ and $\nu = \Var[f(S)]$ be the expected value and the variance of $f(S)$ respectively. For every constant $p \in (0,1)$, and a given parameter $\xi \in (0,1)$, the following holds. 
\begin{enumerate}
    \item There is a $p$-tampering attack $\Adv_\tam$ such that,
$$
\Ex_{\mal{\cS} \gets \adv_\tam(S)}[f(\mal{\cS})] \geq \mu + \frac{p \cdot \nu}{1 + p\cdot \mu - p} - \xi
$$
and given oracle access to $f$ and sampling oracle for $D$, the tampering algorithm $\Tamp^{D,f}_\tam$ of $\Adv_\tam$ could be implemented in time $\poly(|D|\cdot n/\xi)$ where $|D|$ is the bit length of $d \gets D$.
\item  There is a $p$-resetting attack $\Adv_\res$ such that,
$$
\Ex_{\mal{\cS} \gets \adv_\res(S)}[f(\mal{\cS})] \geq \mu + \frac{p\cdot \nu}{1 + p\cdot \mu} - \xi
$$
and given oracle access to $f$ and sampling oracle for $D$, the tampering algorithm $\Tamp^{D,f}_\res$ of $\Adv_\res$ could be implemented in time $\poly(|D|\cdot n/\xi)$ where $|D|$ is the bit length of $d \gets D$.
\end{enumerate}
\end{ourtheorem}
See Section \ref{sec:AttProof} for the full proof of Theorem \ref{thm:main-form}.
In this section, we use Theorem  \ref{thm:main-form} and obtain the following improved attacks in the targeted setting against any learner. In particular, for any fixed $(x,y)=d \gets D$, the following corollary   follows from Theorem \ref{thm:main-form} by letting $f(\cS) = \Ex_{h \gets L(\cS)}[\Loss(h(x),y)]$.
\begin{ourcorollary}[Improved targeted $p$-tampering attacks] \label{cor:Targ}
Given a problem $\problem=(\XX,\YY,\distC,\hypoC,\loss)$ with a bounded loss function $\Loss$, for any distribution $D \in \distC$, test example $(x,y)=d \gets D$, learner $L$, and $n \in \N$,  let $\mu = \error_{D}(d)$  be the average error for $d$, and let,
$$ \nu = \Var_{\cS \gets D^n}\Big[\Ex_{h \gets L(\cS)}[\Loss(h(x),y)]\Big].$$
Then, for any constant $0< p<1$, and any $0<\xi<1$ 
there is a $p$-tampering (resp. $p$-resetting) attack $\Adv_{\tam}$ (resp. $\adv_\res$) that increases the average error by $\frac{p \cdot \nu}{1 + p\cdot \mu - p} - \xi$ (resp. $\frac{p \cdot \nu}{1 + p\cdot \mu }-\xi$). Moreover, if $D$ is polynomial-time samplable and both functions $f,\Loss$ are polynomial-time computable, then  $\Adv_{\tam},\Adv_{\res}$  could be implemented in $\poly(|D|\cdot n/\xi)$ time.
\end{ourcorollary}

\begin{remark}
Even when the average error $\mu = \error_{D}(d)$ is not too small, the variance $\nu$ (as defined in Corollary \ref{cor:Targ}) could be negligible in general. 
However, for natural cases this cannot happen. For example, if the loss function $\Loss(\cdot)$ is Boolean (e.g., $\problem$ is a classification problem) 
and if $L$ is a deterministic learning algorithm, then $\nu = \mu \cdot (1-\mu)$. 
\end{remark}

We now demonstrate the power of $p$-tampering and $p$-resetting attacks on PAC learners by using them to increase the failure probability of deterministic PAC learners.  
\begin{ourcorollary}[$p$-tampering attacks on PAC learners] \label{cor:NonTar}
Given a problem $\problem=(\XX,\YY,\distC,\hypoC,\loss)$, $D \in \distC, n \in \N$, and deterministic learner $L$, suppose,
$$\Pr_{\substack{\cS \gets D^n,~ h = L(\cS)}}[\Risk_\dist( h) \geq \eps] = \delta.$$
Then, there is a $\poly(|D|\cdot n/\eps)$ time $p$-tampering attack $\Adv_{\tam}$ and a $p$-resetting attack $\Adv_\res$ such that,
\begin{align*}
\Pr_{\substack{\cS \gets D^n, {\mal{\cS} \gets \Adv_\tam(\cS)}, h = L(\mal{\cS})}}[\Risk_\dist( h) \geq 0.99 \cdot \eps] &\geq \delta + \frac{p \cdot (\delta-\delta^2)}{1 + p\cdot \delta - p} - \e^{-n} \\
\Pr_{\substack{\cS \gets D^n,  {\mal{\cS} \gets \Adv_\res(\cS)}, h = L(\mal{\cS})}}[\Risk_\dist( h) \geq 0.99 \cdot \eps] &\geq \delta + \frac{p \cdot (\delta-\delta^2)}{1 + p\cdot \delta} - \e^{-n}.
\end{align*}
\end{ourcorollary}

Before proving this we prove a useful proposition.

\begin{proposition}\label{prop:bias-increasing}
The following functions  are increasing for  $\delta\in[0,1]$ and any constant $p\in (0,1)$.
$$\gamma_\tam(\delta)=  \delta + \frac{p\cdot(\delta - \delta^2)}{1+p\cdot \delta -p}, ~~~~~~\gamma_\res(\delta)=  \delta + \frac{p\cdot(\delta - \delta^2)}{1+p\cdot \delta}.$$
\end{proposition}
\begin{proof} The lemma holds because we have, 
$$\frac{\partial \gamma_\tam}{\partial \delta} = \frac{1-p}{(p(\delta-1)+ 1)^2} > 0 \text{ \ and \ } \frac{\partial \gamma_\res}{\partial \delta} = \frac{1+p}{(p\cdot\delta+ 1)^2} > 0\,.$$
\end{proof}
\begin{proof} [Proof of Corollary \ref{cor:NonTar}]
The inefficient versions of the attacks follow from Theorem \ref{thm:main-form} by letting $f(\cS) = 1$ if $\Risk_\dist( h) \geq \eps$ and $f(\cS)=0$ otherwise. When the attacks are supposed to run in polynomial time, we have to approximate $\Risk_\dist( h) $ using oracle access to $D$. Suppose we have access to some oracle $\tilde{f}(.)$ such that,
$$\tilde{f}(\cS)= \begin{cases}
1 & \text{if } \Risk_\dist(L(\cS)) \geq \eps,\\
0 & \text{if } \Risk_\dist(L(\cS)) \leq 0.99\cdot \eps,\\
0 \text{ or } 1 & \text {if } 0.99\cdot\eps \leq \Risk_\dist(L(\cS)) \leq \eps.
\end{cases}
$$
We first show that by using the oracle $\tilde{f}(.)$  instead of $f(\cdot)$, we can achieve the desired bias, and then we will approximate $f(\cdot)$ using oracle access to a sampling oracle for $D$ such that we obtain a simulated oracle for $\tilde{f}(.)$ with probability $1-\e^{-n}$.

If $\tilde{\delta} = \Ex_{\cS \gets D^n}[\tilde{f}(\cS)]$, then Theorem \ref{thm:main-form} shows that given oracle access to $\tilde{f}(.)$, there is a $p$-tampering attack $\Adv_\tam$ and a $p$-resetting attack $\Adv_\res$ that can bias the average of $\tilde{f}$ as,
$$\Ex_{\substack{\cS \gets D^n, {\mal{\cS} \gets \Adv_\tam(\cS)}}}[\tilde{f}(\hat{\cS})]\geq \tilde{\delta} + p\cdot\frac{p\cdot(\tilde{\delta} - \tilde{\delta}^2)}{1+p\cdot \tilde{\delta} -p},~~~~~ \Ex_{\substack{\cS \gets D^n, {\mal{\cS} \gets \Adv_\res(\cS)}}}[\tilde{f}(\hat{\cS})]\geq \tilde{\delta} + p\cdot\frac{p\cdot(\tilde{\delta} - \tilde{\delta}^2)}{1+p\cdot \tilde{\delta}}.  $$
  On the other hand, we know that for all $\cS\in \Supp(D^n), f(\cS) \leq \tilde{f}(\cS).$ 
  Therefore,
 $$\delta = \Ex_{\cS \gets D^n}[f(\cS)] \leq \tilde{\delta}.$$
 We also know that $\tilde{f}(\cS) = 1$ implies that $\Risk(L(\cS)) \geq 0.99\cdot \eps$, thus for any distribution $Z$ defined on $\Supp(D^n)$ we have,
 $$\Ex_{\substack{\mal{\cS} \gets Z}}[\tilde{f}(\hat{\cS})] \leq \Pr_{\substack{\mal{\cS} \gets Z}}[\Risk(L(\hat{\cS})) \geq 0.99\cdot \eps].$$
Combining the above inequalities for the $p$-tampering attack, we get,
\begin{align*}
    \Pr_{\substack{\cS \gets D^n, {\mal{\cS} \gets \Adv_\tam(\cS)}}}[\Risk(L(\hat{\cS})) \geq 0.99\cdot \eps]&\geq \Ex_{\substack{\cS \gets D^n, {\mal{\cS} \gets \Adv_\tam(\cS)}}}[\tilde{f}(\hat{\cS})]\\
    &\geq \hat{\delta} + p\cdot\frac{p\cdot(\tilde{\delta} - \tilde{\delta}^2)}{1+p\cdot \tilde{\delta} -p}\\
    \text{(By Proposition \ref{prop:bias-increasing})~~}& \geq \delta + p\cdot\frac{p\cdot(\delta - \delta^2)}{1+p\cdot \delta -p}.
\end{align*}
Similarly, for the $p$-resetting attack we get,
\begin{align*}
    \Pr_{\substack{\cS \gets D^n, {\mal{\cS} \gets \Adv_\res(\cS)}}}[\Risk(L(\hat{\cS})) \geq 0.99\cdot \eps]&\geq \Ex_{\substack{\cS \gets D^n, {\mal{\cS} \gets \Adv_\res(\cS)}}}[\tilde{f}(\hat{\cS})]\\
    &\geq \hat{\delta} + p\cdot\frac{p\cdot(\tilde{\delta} - \tilde{\delta}^2)}{1+p\cdot \tilde{\delta}}\\
    \text{(By Proposition \ref{prop:bias-increasing})~~}& \geq \delta + p\cdot\frac{p\cdot(\delta - \delta^2)}{1+p\cdot \delta}.
\end{align*}
Now, we show how to obtain an oracle $\tilde{f}(.)$ that provides the properties above with high probability by accessing sampling oracle for $D$. The simulated oracle $\tilde{f}(.)$ works as follows. Given a training set $\cS$, it first performs $L$ on $\cS$ to get the hypothesis $h$. Then it samples $m$ examples $d_1=(x_1,y_1),\dots, d_m=(x_m,y_m)$ from $D^m$, for $m$ to be chosen later, and it computes an ``empirical risk'' $r(h)$ as follows:  $r(h) = \frac{1}{m}\sum_{i=1}^{m} \Loss(h,x_i,y_i)$. If  $r(h) \geq 0.995\time\eps $, $\tilde{f}(\cS)$ outputs $1$, otherwise it outputs $0$. By Hoeffding's inequality, it holds that,
$$\Pr[|r(h)-\Risk_D(h)|\geq 0.005\cdot \eps] \leq 2\cdot \e^{-\frac{m\cdot \eps^2}{20000}}.$$ 
Therefore,
$$\Pr[((r(h)\leq0.995\cdot\eps)\wedge(\Risk_D(h)\geq\eps))\vee((r(h)\geq0.995\cdot\eps)\wedge(\Risk_D(h)\leq 0.99\cdot\eps))]\leq 2\cdot \e^{-\frac{m\cdot \eps^2}{20000}}$$
which means that the oracle $\tilde{f}(.)$ has the required properties with very high probability.
Now, if the original attacker $\Adv_\tam$ or $\Adv_\res$ runs in time $t=\poly(|D|\cdot n/\eps)$, we choose $m=\poly(|D|\cdot n/\eps)$ large enough such that $t \cdot \e^{-\frac{m\cdot \eps^2}{20000}} \leq \e^{-n}$. In particular, we choose $m \geq (n + \ln (2t)) \cdot 20000 / \eps^2$. Therefore, by a union bound, with probability $1-\e^{-n}$, all the queries to $\tilde{f}(\cdot)$ would be within $\pm \eps/200$ of the answer that the ideal oracle $f(\cdot)$ would provide. This concludes the proof of the corollary.
 \end{proof}

\subsection{New  \ensuremath{p}-Tampering and \ensuremath{p}-Resetting Biasing Attacks} \label{sec:AttProof}

In this subsection and Subsection \ref{sec:efficiency} we prove Theorem \ref{thm:main-form}. 
Our 
focus is on describing the relevant tampering algorithms $\Tamp$; 
the general attacks will be defined accordingly. 
(Recall Definition \ref{def:pTamp} and that the $p$-tampering attacker has an internal `tampering' algorithm $\Tamp$ that is executed with independent probability $p$.) 
We  first describe our tampering algorithms in an ideal model where certain parameters of the function $f$ are given for free by an oracle. In Section~\ref{sec:efficiency}, we   get rid of this  assumption by approximating these parameters in polynomial time.
\begin{ourdefinition}[Function $\hat{f}$]\label{defs:biasing} Let $D$ be a distribution, $f \colon \Supp(S) \To \R$ be defined over $D^n$ for some $n\in \N$, and  $\pfix{d}{i}\in \Supp(D)^i$ for some $i \in [n]$. We define the following functions.
\begin{itemize}
  \item  $f_{d_{\leq i}}(\cdot)$ is a function defined as $f_{d_{\leq i}}(d_{\geq i+1}) = f(z)$ where $z=(\pfix{d}{i},d_{\geq i+1}) = (d_1,\dots,d_n)$.
  \item  $\fhat{d}{{ i}}= \Ex_{d_{\geq i+1}\gets D^{n-i}}[f_{d_{\leq i}}(d_{\geq i+1}) ]$. We also use $\mu=\fhat{\es}{}$ to denote $\hat{f}[\pfix{d}{0}]=\Ex[f(S)]$.
 \end{itemize}
\end{ourdefinition}
 The key idea in both of our attacks is to design them (to run in polynomial time) based on oracle access to $\hat{f}$. The point is that $\hat{f}$ could later be  approximated within arbitrarily small $1/\poly(n)$ factors, thus leading to sufficiently close approximations of our attacks.
After describing the `ideal' version of the attacks, we will describe how to make them  efficient by approximating oracle calls to $\hat{f}$.

\paragraph{Changing the range of $f(\cdot)$.} In both of our attacks, we describe our attacks using functions with range $[-1,+1]$. To get the results of Theorem \ref{thm:main-form} we simply need to scale the parameters back appropriately.

\subsubsection{New \ensuremath{p}-Tampering Biasing Attack (Ideal Version)}

Our Ideal \ensuremath{\pTam} attack below, might repeat a loop indefinitely, but as we will see in Section \ref{sec:efficiency}, we can cut this rejection sampling procedure after a large enough polynomial number of rejection trials.
\begin{ourconstruction}[Ideal \ensuremath{\pTam} tampering]\label{cons:ptam}
Let $D$ be an arbitrary distribution and $S \equiv D^n$ for some $n\in N$. Also let $f \colon \Supp(D)^n \To \pmint$ be an arbitrary function.\footnote{As mentioned before,  we describe our attacks using  range $[-1,+1]$, and then we will do the conversion back to $[0,1]$.} For any $i\in [n]$, given a prefix $\pfix{d}{i-1} \in \Supp(D)^{i-1}$,\footnote{Note that here $d_i$ is the `original' untampered value for block $i$, while $d_1,\dots,d_{i-1}$ might be the result of tampering.} \emph{ideal $\pTam$} is a $p$-tampering attack defined as follows. 
\begin{enumerate}
    \item Let $r[\pfix{d}{i}]=\frac{1- \fhat{d}{i}}{3-p-(1-p)\cdot\fhat{d}{i-1}}$.  
    \item With probability $1 - r[\pfix{d}{i}]$ return $d_i$. 
          Otherwise, sample a fresh $d_i \gets D$ and go to step 1. 
\end{enumerate}
\end{ourconstruction}
\begin{ourproposition}
Ideal $\pTam$ attack is well defined. Namely, $r[\pfix{d}{i}] \in [0, 1]$ for all $\pfix{d}{i} \in \Supp(D)^i$. 
\end{ourproposition}
\begin{proof}
Both $\fhat{d}{i},\fhat{d}{i-1}$ are in $[-1, 1] $. Therefore $0\leq1- \fhat{d}{i}\leq 2$ and $3-p-(1-p)\cdot\fhat{d}{i-1}\geq2$ which implies $0\leq r[\pfix{d}{i}]\leq 1$.
\end{proof}
In the following, let $\adv_\tam$ be the $p$-tampering adversary using tampering algorithm Ideal  $\pTam$.\footnote{Therefore, $\Adv_D$, inductively runs $\pTam$ over the current sequence with probability $p$. See Definition \ref{def:pTamp}.}
\begin{ourclaim}\label{clm:ptam_dist}
Let $\mal{S}=(\mal{D}_1,\dots,\mal{D}_n)$ be the joint distribution after $\adv_\tam$ attack is performed on $S \equiv D^n$ using ideal $\pTam$ tampering algorithm. For every prefix $\pfix{d}{i} \in \Supp(D)^i$ we have,
$$\frac{\Pr[\mal{D}_i=d_i \mid \pfix{d}{i-1}]}{\Pr[D=d_i]} = \frac{2 -p \cdot (1-\fhat{d}{i})}{2-p\cdot (1- \fhat{d}{i-1})}.$$
\end{ourclaim}
\begin{proof}
During its execution, ideal $\pTam$ keeps sampling examples and rejecting them until a sample is accepted. For $\ell \in \N$ we define $\Rej_\ell$ to be the event that is true if the $\ell$'th sample in the tampering algorithm is rejected, conditioned on reaching the $\ell$th sample. We have,
\begin{align*}
\Pr[\Rej_\ell] &= \sum_{d_i} \Pr[D=d_i]\cdot \left(\frac{1-\fhat{d}{i}}{3-p-(1-p)\cdot\fhat{d}{i-1}}\right)\\&= \frac{\sum_{d_i} \Pr[D=d_i]\cdot(1-\fhat{d}{i})}{3-p-(1-p)\cdot\fhat{d}{i-1}} = \frac{1-\fhat{d}{i-1}}{3-p-(1-p)\cdot\fhat{d}{i-1}}.
\end{align*}
Let  $c[\pfix{d}{i-1}] = \frac{1-\fhat{d}{i-1}}{3-p-(1-p)\cdot\fhat{d}{i-1}}$. Then we have,
\begin{align*}
\frac{\Pr[\mal{D}_i=d_i \mid \pfix{d}{i-1}]}{\Pr[D=d_i]} &= 1-p+ p\cdot\left( \sum_{j=0}^{\infty}\left(1-r[\pfix{d}{i}]\right)\cdot \prod_{\ell=1}^{j}\Pr[\Rej_\ell] \right) \\
&=1-p + p\cdot\left( \sum_{j=0}^{\infty}(1-r[\pfix{d}{i}])\cdot c[\pfix{d}{i-1}]^j \right)\\
&=1-p + p\cdot\left(\frac{1-r[\pfix{d}{i}]}{1-c[\pfix{d}{i-1}]}\right) = \frac{2-p + p\cdot \fhat{d}{i}}{2-p + p\cdot \fhat{d}{i-1}}. 
\end{align*}
\end{proof}

\remove{
We have 
\begin{eqnarray}
1-p + p\cdot\left(\frac{1-r[\pfix{d}{i}]}{1-c[\pfix{d}{i-1}]}\right) 
& = & 1-p + p\cdot\left(\frac{1-r[\pfix{d}{i}]}{1-\frac{1-\fhat{d}{i-1}}{3-p-(1-p)\cdot\fhat{d}{i-1}}}\right) \nonumber\\
& = & 1-p + p\cdot\left(\frac{(1-r[\pfix{d}{i}])\cdot (3-p-(1-p)\cdot\fhat{d}{i-1})}{3-p-(1-p)\cdot\fhat{d}{i-1} - 1 + \fhat{d}{i-1}}\right) \nonumber\\
& = & 1-p + p\cdot\left(\frac{(1-r[\pfix{d}{i}])\cdot (3-p-(1-p)\cdot\fhat{d}{i-1})}{2-p + p\cdot\fhat{d}{i-1}}\right) \nonumber\\
& = & 
\frac{2-p + p\cdot\fhat{d}{i-1} - 2p+p^2 - p^2\cdot\fhat{d}{i-1}}{2-p + p\cdot\fhat{d}{i-1}} \nonumber\\
& & + \frac{p\cdot(1-r[\pfix{d}{i}])\cdot (3-p-(1-p)\cdot\fhat{d}{i-1})}{2-p + p\cdot\fhat{d}{i-1}}\nonumber \\
& = & 
\frac{2-p + p\cdot\fhat{d}{i-1} - 2p+p^2 - p^2\cdot\fhat{d}{i-1}}{2-p + p\cdot\fhat{d}{i-1}} \nonumber\\
& & + \frac{p\cdot (3-p-(1-p)\cdot\fhat{d}{i-1})}{2-p + p\cdot\fhat{d}{i-1}} \nonumber\\
& & - \frac{p\cdot \frac{1- \fhat{d}{i}}{3-p-(1-p)\cdot\fhat{d}{i-1}}\cdot (3-p-(1-p)\cdot\fhat{d}{i-1})}{2-p + p\cdot\fhat{d}{i-1}} \nonumber\\
& = & 
\frac{2-p + p\cdot\fhat{d}{i-1} - 2p+p^2 - p^2\cdot\fhat{d}{i-1}}{2-p + p\cdot\fhat{d}{i-1}} \nonumber\\
& & + \frac{p\cdot (3-p-(1-p)\cdot\fhat{d}{i-1})}{2-p + p\cdot\fhat{d}{i-1}} \nonumber\\
& & - \frac{p - p\fhat{d}{i}}{2-p + p\cdot\fhat{d}{i-1}} \nonumber\\
& = & 
\frac{2-p + p\cdot\fhat{d}{i}}{2-p + p\cdot\fhat{d}{i-1}} + Q\nonumber
\end{eqnarray}
where
\begin{eqnarray}
Q & = & \frac{p\cdot\fhat{d}{i-1} - 2p+p^2 - p^2\cdot\fhat{d}{i-1} 
+ p\cdot (3-p-(1-p)\cdot\fhat{d}{i-1})
- p
}{2-p + p\cdot\fhat{d}{i-1}} \nonumber\\
& = & \frac{p\cdot\fhat{d}{i-1} - 2p+\cancel{p^2} - p^2\cdot\fhat{d}{i-1} 
+ 3p-\cancel{p^2}-p\cdot\fhat{d}{i-1} + p^2\cdot\fhat{d}{i-1}
- p
}{2-p + p\cdot\fhat{d}{i-1}} \nonumber\\
& = & 0 \nonumber
\end{eqnarray}
}

The next corollary follows from Claim \ref{clm:ptam_dist} and induction.
(Recall that $\mu=\fhat{\es}{} = \hat{f}[\pfix{d}{0}]=\Ex[f(S)]$.)
\begin{ourcorollary}
By applying the attack $\adv_\tam$ based on the ideal $\pTam$ tampering algorithm, the distribution after the attack would be as follows,
$$\Pr[\mal{S}=z] = \frac{2-p + p\cdot f(z)}{2-p + p\cdot\mu}\cdot \Pr[S=z].$$
\end{ourcorollary}
\begin{ourcorollary}\label{cor:ptambias}
The $p$-tampering attack $\adv_\tam$ (based on the ideal $\pTam$ tampering algorithm) biases $f(\cdot)$ by $\frac{p\cdot\nu}{2-p + p\cdot\mu}$ 
    where $\mu=\Ex[f(S)], \nu = \Var[f(S)]$.
\end{ourcorollary}
\begin{proof}
It holds that $\Ex[f(\mal{S})]$ is equal to
\begin{align*}
    &\sum_{z\in \Supp(D)^n}\Pr[\mal{S}=z]\cdot f(z) = \sum_{z\in \Supp(D)^n}\frac{2-p + p\cdot f(z)}{2-p +p\cdot \mu}\cdot \Pr[S=z] \cdot f(z) \\
&= \frac{2-p}{2-p + p\cdot\mu}\cdot \left(\sum_{z\in \Supp(D)^n}\Pr[S=z]\cdot f(z)\right) + \frac{p}{2-p + p\cdot\mu}\cdot \left(\sum_{z\in \Supp(D)^n}\Pr[S=z]\cdot f(z)^2\right) \\
    &= \frac{(2-p)\cdot \mu}{2-p + p\cdot\mu} + \frac{p\cdot (\nu + \mu^2)}{2-p + p\cdot\mu}= \mu + \frac{p\cdot\nu}{2-p + p\cdot\mu}.  
\end{align*}
\end{proof}
\begin{ourcorollary}
For any $S \equiv D^n$ and any function $f\colon \Supp(D^n) \to [0,1]$, there is a $p$-tampering attack that given oracle access to $\hat{f}(\cdot)$ and a sampling oracle for $D$, it biases the expected value of $f$ by $\frac{p\cdot\nu}{1-p + p\cdot\mu}$ 
    where $\mu=\Ex[f(S)], \nu = \Var[f(S)]$. 
\end{ourcorollary}
\begin{proof}
Consider another function $f' = 2\cdot f -1$. The range of $f'$ is now $\pmint$ and we have $\mu' = \Ex[f'(S)] = 2\cdot\mu-1$ and $\nu' = \Var[f'(S)] = 4\cdot \nu$. By Corollary \ref{cor:ptambias},  the $p$-tampering attack $A_\tam$ biases $f'$ by $\frac{p\cdot \nu'}{2-p+p\cdot \mu'}$. Let $\mal{S}$ be the tampered distribution after performing $A_\tam$ on function $f'$ and $S$. We have,
$$\Ex[f'(\mal{S})] \geq \mu' + \frac{p\cdot \nu'}{2-p+p\cdot \mu'}.$$ Therefore we have,
$$\Ex[f(\mal{S})] = \frac{\Ex[f'(\mal{S})]+1}{2} \geq \frac{\mu'+1}{2} + \frac{p\cdot \nu'}{2\cdot(2-p+p\cdot \mu')} =  \mu + \frac{p\cdot\nu}{1-p+p\cdot \mu}.$$
\end{proof}
\subsubsection{New \ensuremath{p}-Resetting Biasing Attack (Ideal Version)}
\begin{ourconstruction}[Ideal $\pRes$] \label{cons:pres}
Let $D$ be an arbitrary distribution and $S \equiv D^n$ for some $n\in N$. Also let $f \colon \Supp(D)^n \To \pmint$ be an arbitrary function.\footnote{As mentioned before,  we describe our attacks using  range $[-1,+1]$, and then we will do the conversion back to $[0,1]$.} For any $i\in [n]$, and given a prefix $\pfix{d}{i-1} \in \Supp(D)^{i-1}$, the  $\pRes$ tampering algorithm works as follows. 
\begin{enumerate}
    \item Let $r[\pfix{d}{i}]= \frac{1- \fhat{d}{i}}{2+p\cdot(1 +\fhat{d}{i-1})}$.
    \item With probability $1 - r[\pfix{d}{i}]$ output the given $d_i$. 
    \item Otherwise sample $d'_i \gets D$ (i.e., `reset' $d_i$) and return $d'_i$.
\end{enumerate}
\end{ourconstruction}
\begin{ourproposition}
Ideal $\pRes$  algorithm is well defined. Namely, $r[\pfix{d}{i}]\in[0, 1]$ for all $\pfix{d}{i}\in \Supp(D)^i$.
\end{ourproposition}
\begin{proof}
We have $\fhat{d}{i} \in [-1, +1] $ and  $\fhat{d}{i-1} \in [-1, +1]$. Therefore $0\leq1- \fhat{d}{i}\leq 2$ and $2+p\cdot(1 +\fhat{d}{i-1})\geq2$ which implies $0\leq r[\pfix{d}{i}]\leq1$.
\end{proof}
In the following let $\adv_\res$ be the $p$-tampering adversary using ideal $\pRes$. (See Definition \ref{def:pTamp}.)
\begin{ourclaim}\label{clm:pres_dist}
Let $\mal{S}=(\mal{D}_1,\dots,\mal{D}_n)$ be the distribution after the attack $\adv_\res$ (using ideal $\pRes$ tampering algorithm) is performed on $S \equiv D^n$. For all  $\pfix{d}{i} \in \Supp(D)^i$ it holds that,
$$\frac{\Pr[\mal{D}_i=d_i \mid \pfix{d}{i-1}]}{\Pr[D=d_i]} = \frac{2 +p \cdot (1+\fhat{d}{i})}{2+p\cdot(1+ \fhat{d}{i-1})}.$$
\end{ourclaim}
\begin{proof}
We define $\Rej_1$ to be the event that is true if the given sample is rejected. We have,
\begin{align*}
\Pr[\Rej_1] &= \sum_{d_i} \Pr[D=d_i]\cdot \left(\frac{1-\fhat{d}{i}}{2+p\cdot(1 +\fhat{d}{i-1})}\right) \\
&= \frac{\sum_{d_i} \Pr[D=d_i]\cdot(1-\fhat{d}{i})}{2+p\cdot(1 +\fhat{d}{i-1})} = \frac{1-\fhat{d}{i-1}}{2+p\cdot(1 +\fhat{d}{i-1})}.
\end{align*}
Therefore, we conclude that,
\begin{align*}
\frac{\Pr[\mal{D}_i=d_i\mid\pfix{d}{i-1}]}{\Pr[D=d_i]} &= 1-p+ p\cdot(1-r[\pfix{d}{i}] + \Pr[\Rej_1])\\
&=1-p + p\cdot\left(1 + \frac{\fhat{d}{i}-\fhat{d}{i-1}}{2+p\cdot(1 +\fhat{d}{i-1})} \right)\\
&=1+ p\cdot\left(\frac{\fhat{d}{i}-\fhat{d}{i-1}}{2+p\cdot(1+\fhat{d}{i-1})}\right) =\frac{2+p\cdot(1+\fhat{d}{i})}{2+p\cdot(1+\fhat{d}{i-1})}. 
\end{align*}
\end{proof}
The next corollary follows from Claim \ref{clm:pres_dist} and induction.
(Recall that $\mu=\fhat{\es}{} = \hat{f}[\pfix{d}{0}]=\Ex[f(S)]$.)
\begin{ourcorollary}
By applying attack $\adv_\res$ (using ideal $\pRes$), the distribution after the attack is,
$$\Pr[\mal{S}=z] = \frac{2+p + p\cdot f(z)}{2+p + p\cdot\mu}\cdot \Pr[S=z].$$
\end{ourcorollary}
\begin{ourcorollary}\label{cor:presbias}
The $p$-resetting attack $\adv_\res$ (using ideal $\pRes$) biases the function by $\frac{p\cdot\nu}{2+p + p\cdot\mu}$ 
    where $\mu=\Ex[f(S)], \nu = \Var[f(S)]$.
\end{ourcorollary}
\begin{proof}
It holds that $\mal{\mu}=\Ex[f(\mal{S})]$ is equal to
\begin{align*}
    & \sum_{z\in \Supp(D)^n}\Pr[\mal{S}=z]\cdot f(z) = \sum_{z\in \Supp(D)^n}\frac{2+p + p\cdot f(z)}{2+p +p\cdot \mu}\cdot \Pr[S=z] \cdot f(z)\\
    &= \frac{2+p}{2+p + p\cdot\mu}\cdot \left(\sum_{z\in \Supp(D)^n}\Pr[S=z]\cdot f(z)\right) + \frac{p}{2+p + p\cdot\mu}\cdot \left(\sum_{z\in \Supp(D)^n}\Pr[S=z]\cdot f(z)^2\right)\\
    &= \frac{(2+p)\cdot \mu}{2+p + p\cdot\mu} + \frac{p\cdot (\nu + \mu^2)}{2+p + p\cdot\mu}= \mu + \frac{p\cdot\nu}{2+p + p\cdot\mu}. 
\end{align*}
\end{proof}
\begin{ourcorollary}
For $S \equiv D^n$ and any  $f\colon \Supp(S) \to [0,1]$ there exist a $p$-resetting attack that, given oracle access to $\hat{f}$ and a sampling oracle for $D$, it biases $f$ by $\frac{p\cdot\nu}{1+ p\cdot\mu}$ 
    where $\mu=\Ex[f(S)], \nu = \Var[f(S)]$. 
\end{ourcorollary}
\begin{proof}
Consider another function $f' = 2\cdot f -1$. Now, the range of $f'$ is $\pmint$, and we have $\mu' = \Ex[f'(S)] = 2\cdot\mu-1$ and $\nu' = \Var[f'(S)] = 4\cdot \nu$. By Corollary \ref{cor:presbias}, the $p$-resetting attack $A_\res$ biases $f'$ by $\frac{p\cdot \nu'}{2-p+p\cdot \mu'}$. Let $\mal{S}$ be the tampered distribution after performing $A_\tam$ on function $f'$ and $S$. We have,
$$\Ex[f'(\mal{S})] \geq \mu' + \frac{p\cdot \nu'}{2+p+p\cdot \mu'}.$$ Therefore we have,
$$\Ex[f(\mal{S})] = \frac{\Ex[f'(\mal{S})]+1}{2} \geq \frac{\mu'+1}{2} + \frac{p\cdot \nu'}{2\cdot(2+p+p\cdot \mu')} =  \mu + \frac{p\cdot\nu}{1+p\cdot \mu}.$$
\end{proof}
%



\subsection{Approximating the Ideal Attacks in Polynomial Time} \label{sec:efficiency}
In this subsection, we describe the efficient version of the attacks of Theorem \ref{thm:main-form} and prove their properties. We first describe the efficient version of our $p$-resetting attack, where achieving efficiency is indeed simpler. We then go over the efficient variant of our $p$-tampering attack. In both cases, we describe the modifications needed for the \emph{tampering algorithms} and it is assumed that such tampering algorithms are used by the main efficient attackers (see Definition \ref{def:pTamp}).

We start by  approximating in polynomial time our Ideal \ensuremath{p}-resetting attack, as it is simpler to argue about the polynomial-time version of this attack. We will then use lemmas and ideas that we develop along the way to also make our 1st Ideal \ensuremath{p}-tampering attacker also polynomial time.

\subsubsection{Polynomial-time Variant of the Ideal \ensuremath{p}-Resetting Biasing Attack}
The $p$-resetting attack of  Construction \ref{cons:pres} is not polynomial-time since it needs oracle access to the idealized oracle providing partial averages.  In general, we can not compute such averages exactly in polynomial time, however in order to make those attacks polynomial-time, we can rely on \emph{approximating} the partial averages and consequently the  corresponding rejection probabilities. To get the polynomial-time version of the attack of Construction \ref{cons:pres} we can pursue the following idea. For every prefix $\pfix{d}{i}$, the polynomial-time attacker first approximates the partial average $\fhat{d}{i}$ by sampling a sufficiently large polynomial number of random continuations $\pfix{d^{(1)}}{n-i},\dots \pfix{d^{(\ell)}}{n-i}$ and getting the average $\Ex_{j\in [\ell]} [f(\pfix{d}{i},\pfix{d^{(j)}}{n-i}]$ as an approximation for the partial average. By Hoeffding inequality, this average is a good approximation of $\fhat{d}{i}$ with exponentially high probability.  
Consequently, the rejection probabilities can be approximated well making the final distributions statistically close to the distribution of the ideal attack, meaning that the amount of bias is close to the ideal bias as well.

Now we formalize the ideas above.

\begin{ourdefinition}[Semi-ideal oracle \ensuremath{\tilde{f}[\cdot]}]
For distribution $D$, if for all  $d_{\leq i} \in \Supp(D)^i$ we have $\tilde{f}_\xi[d_{\leq i}] \in \fhat{d}{i} \pm \xi$, then, we call $\tilde{f}_\xi[\cdot]$  an $\xi$-approximation of $\hat{f}[\cdot]$. For simplicity, and when it is clear from the context, we simply write $\tilde{f}[\cdot]$ and call it a \emph{semi-ideal} oracle.
\end{ourdefinition}

The following lemma immediately follows from the Hoeffding inequality.
\begin{ourlemma}[Approximating \ensuremath{\hat{f}[\cdot]} in polynomial-time] \label{lemma:approx_fhat} Consider an algorithm that on inputs $\pfix{d}{i}$ and $\xi$ performs as follows where $\ell={-10\ln(\xi/2)}/{\xi^2}$. 
\begin{enumerate}
    \item Sample $(\pfix{d}{n-i}^1,\dots, \pfix{d}{n-i}^\ell) \gets (D^{n-i+1})^\ell$.
    \item Output $\afhat{d}{i} = \Ex_{j \in [\ell]}{f(\pfix{d}{i}, \pfix{d}{n-i}^j)}$.
\end{enumerate}
Then it holds that $\Pr[|\afhat{d}{i} - \fhat{d}{i}|\geq \xi] \leq \xi$.
\end{ourlemma}

The above lemma implies that if $f$ is polynomial-time computable and $D$ is polynomial-time samplable, any $q$-query algorithm can approximate the semi-ideal oracle $\tilde{f}[\cdot]$ in time $\poly(q\cdot n/\xi)$ and total error (of failing in one of the queries) by at most $\xi$. Based on this  approximation of $\tilde{f}[\cdot]$, we now describe our polynomial-time version of the Ideal $\pRes$ attack in the semi-ideal oracle model of $\tilde{f}[\cdot]$, by essentially using the semi-ideal oracle $\tilde{f}[\cdot]$ instead of the ideal oracle $\hat{f}[\cdot]$.

\begin{ourconstruction}[Polynomial-time $\pRes$] \label{cons:epres}
Polynomial-time $\pRes$ is the same as ideal  $\pRes$ of Construction \ref{cons:pres} but it calls the semi-ideal oracle $\tilde{f_\xi}[\cdot]$ instead of the ideal oracle $\hat{f}[\cdot]$.
\end{ourconstruction}

In the following we analyze the bias achieved by the the polynomial-time variant of the $\pRes$ algorithm. We simply pretend that all the queries to the semi-ideal oracle are within $\pm \xi$ approximation of the ideal oracle, knowing that the error of $\xi$-approximating all of the queries is itself at most $\xi$ and can affect the average also by at most $O(\xi)$.
First we show that the rejection probabilities are approximated well.

\begin{ourlemma} \label{lem:approxR}
Let $0 < p < 1$, $0 < \xi < 1$, $\alpha, \beta \in [-\xi, \xi]$, and  $\fhat{d}{i-1}, \fhat{d}{i}, \afhat{d}{i-1}, \afhat{d}{i} \in [0, 1]$ such that 
$\afhat{d}{i-1} = \fhat{d}{i-1} + \alpha$
and
$\afhat{d}{i} = \fhat{d}{i} + \beta$.
Let $r[.]$ and $\tilde{r}[.]$ respectively be the rejection probabilities of the Ideal and Polynomial-time $\pRes$. Then, for every $\pfix{d}{i} \in \Supp(D)^i$,
$|r[\pfix{d}{i}]-\tilde{r}[\pfix{d}{i}]| \leq O(\xi)$.
\end{ourlemma}
\begin{proof}
\remove{Let $p' \in p \pm \xi, q' \in q \pm \xi$ for $ p,q,p',q' \in (0,1)$. 
We first show that $\frac{p'}{1+q'} \in \frac{p}{1+q}\pm O(\xi)$. 
\begin{align*}
    \left|\frac{p'}{1+q'}-\frac{p}{1+q}\right|= \left|\frac{p'-p + p'\cdot q -p\cdot q'}{(1+q)\cdot (1+q')}\right|
    \leq \left|p'-p + p'\cdot (q-q') + q'\cdot (p'-p)\right| \leq 3\cdot\xi  
\end{align*}
Now using this general statement we conclude that $|r[\pfix{d}{i}]-\tilde{r}[\pfix{d}{i}]| \leq 3\cdot\xi$.
}
We have, 
\remove{
\begin{align*}
    \left|r[\pfix{d}{i}]-\tilde{r}[\pfix{d}{i}]\right|&= \left|\frac{1- \fhat{d}{i})}{2+p\cdot(1 +\fhat{d}{i})} - \frac{1- \afhat{d}{i}}{2+p\cdot(1 +\afhat{d}{i})}\right|\\
    &= \left|\frac{(2p+2)(\afhat{d}{i}-\fhat{d}{i})}{(2+p\cdot(1 +\fhat{d}{i}))\cdot(2+p\cdot(1 +\afhat{d}{i}))}\right|\\
    &\leq \left|\frac{\cancel{2}(1+p)(\afhat{d}{i}-\fhat{d}{i})}{\cancel{2}\cdot 2}\right|\\
    &\leq \left|\afhat{d}{i}-\fhat{d}{i}\right| \\
    &\leq \xi.\qedhere
\end{align*}
}
\begin{align*}
    \left|r[\pfix{d}{i}]-\tilde{r}[\pfix{d}{i}]\right|
    &= \left|\frac{1- \fhat{d}{i})}{2+p\cdot(1 +\fhat{d}{i-1})} - \frac{1- \afhat{d}{i}}{2+p\cdot(1 +\afhat{d}{i-1})}\right| \,,
\end{align*}
where we can compute the following for the right hand side,  
\begin{align*}
    &= \left|\frac{ (2+p)(\afhat{d}{i} - \fhat{d}{i}) + p \cdot (\afhat{d}{i-1} - \fhat{d}{i-1}) + p \cdot (\fhat{d}{i-1}\afhat{d}{i} - \afhat{d}{i-1}\fhat{d}{i}) }{(2+p\cdot(1 +\fhat{d}{i-1}))\cdot(2+p\cdot(1 +\afhat{d}{i-1}))}\right|\\
    &\leq \frac{\left|(2+p)(\afhat{d}{i} - \fhat{d}{i})\right| + \left|p \cdot (\afhat{d}{i-1} - \fhat{d}{i-1})\right| + \left|p \cdot (\fhat{d}{i-1}\afhat{d}{i} - \afhat{d}{i-1}\fhat{d}{i})\right|}{2\cdot 2} \\
    &\leq \frac{(2 + p)\xi + p\xi + p\left|\fhat{d}{i-1}(\fhat{d}{i}+\beta) - (\fhat{d}{i-1}+\alpha)\fhat{d}{i}\right|}{4} \\
    &= \frac{2\xi + 2p\xi + p\left|\beta\fhat{d}{i-1} - \alpha\fhat{d}{i}\right|}{4} \le \frac{2\xi + 2p\xi + p \cdot (|\beta| + |-\alpha|)}{4} \leq 3\xi/2 \,. \qedhere
\end{align*}
\end{proof}


Now we want to argue that when we approximate the $p$-resetting tampering algorithm's rejection probabilities as proved in Lemma \ref{lem:approxR}, it leads to `close probabilities' of sampling final outputs. We prove the following general lemma that will be also useful for the case of Polynomial-time $\pTam$ attack. For the case of $p$-resetting, we only need the special case of $k=1$.

\paragraph{Notation.} For $p \in [0,1]$ and distributions $X,Y$, by $Z\equiv (1-p)X + pY$ we denote the distribution $Z$ in which we sample from $X$ with probability $1-p$, and otherwise (i.e., with probability $p$) we sample from $Y$.

\begin{ourdefinition}[$(p,k,\rho)$-variations] For any distribution $D$,  function  $\rho\colon \Supp(D) \to [0,1]$, and  $k\in \N$, the $(p,k,\rho)$-variation of $D$ is  $\pkrvar \equiv (1-p)D + pZ$, where $Z$ is defined as follows.
\begin{enumerate}
    \item Sample $(d_1,\dots,d_k) \gets D^k$.
    \item For $i\in\{1, \ldots, k\}$, go sequentially over $d_1,\dots,d_k$, 
          and with probability $\rho[d_i]$ 
          return $d_i$ and exit.
    \item If nothing was returned after reading all the $k$ samples, return a fresh sample $d_{k+1} \gets D$.
\end{enumerate}
\end{ourdefinition}

\begin{ourlemma} [Implication of approximating rejection probabilities]\label{lemma:ImplicationOfApprox}
Let $D$ be a distribution and $\rho:\Supp(D) \to [0,1]$ and $\rho': \Supp(D) \to [0,1]$ be two functions such that $\forall d \in \Supp(D), |\rho(d)-\rho'(d)|\leq \xi$. Then, for every $k\in \N$ and every $d \in \Supp(D)$, it holds that,
 $$ \left|\ln\left(\frac{\Pr[\pkrvar = d]}{\Pr[\pkrvarp=d]}\right)\right|\leq \frac{p}{1-p}\cdot(k^2+k)\cdot\xi.$$
\end{ourlemma}

Before proving the lemma above, we note that it indeed implies that the \emph{max divergence} \citep{boosting2010} of $\pkrvar$ and $\pkrvarp$ is at most $O(k^2 \cdot\xi)$.

\begin{proof}
Let $a = \Ex_{d\gets D} [\rho(d)]$ and $a' = \Ex_{d\gets D} [\rho'(d)]$. We have,
\begin{align*}
    \frac{\Pr[\pkrvar=d]}{\Pr[D=d]} = (1-p) + p\cdot((1-a)^k + \sum_{i\in [k-1]} \rho(d) \cdot (1-a)^i).
\end{align*}
With a similar calculation for $\Pr[\pkrvarp=d]$ we get,
\begin{align*}
    \frac{\Pr[\pkrvar=d]}{\Pr[\pkrvarp=d]} &= \frac{(1-p)+ p\cdot((1-a)^k + \sum_{i\in [k-1]} \rho(d) \cdot (1-a)^i)}{(1-p)+ p\cdot((1-a')^k + \sum_{i\in [k-1]} \rho(d) \cdot (1-a')^i)}\\
    &= 1 + \frac{p\cdot((1-a)^k - (1-a')^k + \sum_{i\in [k-1]} \rho(d)\cdot (1-a)^i  - \rho'(d)\cdot (1-a')^i)}{(1-p)+ p\cdot((1-a')^k + \sum_{i\in [k-1]} \rho(d) \cdot (1-a')^i)}\\
    &\leq 1 +\frac{p\cdot(k\cdot\xi + \sum_{i\in [k-1]} (2i+1)\cdot\xi)}{1-p}\\
    &=1 + \frac{p}{1-p}(k^2+k)\cdot  \xi\\
    &\leq \e^{\frac{p}{1-p}(k^2+k)\cdot \xi}.
\end{align*}
Similarly, we have $\frac{\Pr[\pkrvarp=d]}{\Pr[\pkrvar=d]}\leq \e^{\frac{p}{1-p}(k^2+k)\xi}$ which implies that,
$$\left|\ln\left(\frac{\Pr[\pkrvar = d]}{\Pr[\pkrvarp=d]}\right)\right|\leq \frac{p}{1-p}\cdot(k^2+k)\cdot\xi.$$
\end{proof}

The following lemma states that the expected values of a function over two  distributions that are `close' (under max divergence) are indeed close real numbers.
\begin{ourlemma}\label{lemma:closeAverage}
Let $X=(X_1,\dots,X_n)$ and $Y=(Y_1,\dots,Y_n)$ be two joint distributions such that $\Supp(X) = \Supp(Y)$ and for every prefix $\pfix{x}{i}$ such that $\Pr[X_i = x_i \mid \pfix{x}{i-1}] > 0$, we have,
$$\left|\ln\left(\frac{\Pr[X_i = x_i \mid \pfix{x}{i-1}]}{\Pr[Y_i = x_i \mid \pfix{x}{i-1}]}\right)\right|\leq \xi.$$ 
Then, for any function $f\colon \Supp(X) \to [-1,+1]$ we have,
$$\Ex[f(X)]  \geq \Ex[f(Y)] -\e^{\xi\cdot n} +1.$$
\end{ourlemma}
\begin{proof} 
First, we  note that for every $x\in \Supp(X)$ it holds that,
$$\left|\ln\left(\frac{\Pr[X=x]}{\Pr[Y=x]}\right)\right| = \left|\sum_{i\in [n]}\ln\left(\frac{\Pr[X_i = x_i \mid \pfix{x}{i-1}]}{\Pr[Y_i = x_i \mid \pfix{x}{i-1}]}\right)\right| \leq n\cdot\xi.$$
Now for the difference $\Ex[f(Y)] - \Ex[f(X)]$ we have, 
\begin{align*}
    & \sum_{x \in \Supp(X)} (\Pr[Y=x] - \Pr[X=x])\cdot f(x)\\
    &\leq \sum_{x \in \Supp(X)} \left|(\Pr[Y=x] - \Pr[X=x])\cdot f(x)\right|\\
    &\leq \sum_{x \in \Supp(X)} \left|\min(\Pr[X=x], \Pr[Y=x])\cdot\left(\frac{\max(\Pr[X=x], \Pr[Y=x])}{\min(\Pr[X=x], \Pr[Y=x])}-1\right)\cdot f(x)\right|\\
    &\leq (\e^{n\cdot \xi}-1)\cdot\sum_{x \in \Supp(X)} \left|\min(\Pr[X=x], \Pr[Y=x])\cdot f(x)\right|\\
    &\leq \e^{n\cdot \xi}-1.\qedhere
\end{align*}
\end{proof}

\paragraph{Putting things together.} 
Now we show how to choose the parameters of the Polynomial-time $\pRes$. 
Suppose $\xi'$ is the parameter of Theorem \ref{thm:main-form}. 
If we choose $\xi$ as the parameter of our attack we can bound the final bias as follows. 
Firstly, if the approximation algorithm of Lemma \ref{lemma:approx_fhat} gives us a semi-ideal oracle $\tilde{f_\xi}[.]$, then based on Lemma \ref{lem:approxR} we can approximate the rejection probabilities with error at most $O(\xi)$. 
Then based on  Lemma \ref{lemma:ImplicationOfApprox} the attack $\Adv_\res$ that uses efficient $\pRes$ generates a distribution that is $O(\frac{p}{1-p}\cdot\xi)$-close\footnote{Since we are assuming $p<1$ is constant $O(\frac{p}{1-p}\cdot\xi)$ simply means $O(\xi)$.}
to the distribution of the attack $\Adv_\res$ that uses ideal $\pRes$. 

Now we can use Lemma \ref{lemma:closeAverage} (for $k=1$) to argue that the bias achieved by the efficient adversary is $(\e^{O(n\cdot \xi \cdot  \frac{p}{1-p})}-1)$-close to the bias achieved by the ideal adversary. 
Also note that, if the approximation algorithm fails to provide a semi-ideal oracle for all queries, 
then the bias of the efficient attack is at least $-2$ because the function range is $[-1,+1]$. 
However, the probability of this event is bounded by $O(n\cdot\xi)$ 
because adversary needs at most $2n$ number of queries to $\tilde{f}$. 
Therefore, the difference of bias of the efficient and the ideal adversary is at most $O(n\cdot \xi) + \e^{O(n \cdot \xi\cdot \frac{p}{1-p})}-1$ which is at most $O(n\cdot \xi + n\cdot \xi\cdot \frac{p}{1-p})$ if the exponent in $\e^{O(n\cdot \xi\cdot \frac{p}{1-p})}$ is at most $1$. 
As a result, if we choose $\xi = o \left( \xi'/(n\cdot  \frac{p} {1-p})\right) = o\left(\xi' \cdot (1-p) / (n\cdot p)\right)$, we can indeed guarantee that the bias of the efficient adversary is $\xi'$-close to bias of ideal adversary.

\subsubsection{Polynomial-time Variant of our \ensuremath{p}-Tampering Biasing Attack}
Building upon the ideas developed above to make our Ideal $\pRes$ tampering algorithm polynomial time, here we focus on our Ideal $\pTam$ attack. We start by describing a variant of the original attack of Construction~\ref{cons:ptam} where we cut the rejection sampling procedure after $k$ iterations.

\begin{ourconstruction}[Ideal $k$-cut $\pTam$]\label{const:IdealKcut}
Ideal $k$-cut $\pTam$ is the same as ideal $\pTam$ of Construction \ref{cons:ptam} but it is forced to stop and return a fresh sample if the first $k$ samples were rejected.
\end{ourconstruction}
Now we show that the new modified attack of Construction \ref{const:IdealKcut} will lead to a
 close distribution compared to the original attack of Construction \ref{cons:ptam}.
\begin{ourlemma} \label{lemma:kcut-dist}
Let $\mal{S}=(\mal{D}_1,\dots,\mal{D}_n)$ be the joint distribution after $\adv_\tam$ attack is performed on $S \equiv D^n$ using ideal $\pTam$ tampering algorithm. Also, let $\mal{S}'=(\mal{D}'_1,\dots,\mal{D}'_n)$  be the joint distribution after $\adv_\tam$ attack is performed on $S$ using Ideal $k$-cut $\pTam$ tampering algorithm. For every prefix $\pfix{d}{i} \in \Supp(D)^i$, 
$$\left|\ln\left(\frac{\Pr[\mal{D}_i=d_i \mid \pfix{d}{i-1}]}{\Pr[\mal{D}'_i=d_i|\pfix{d}{i-1}]}\right)\right| \leq \frac{p}{(1-p)^2\cdot (2-p)^{k-1}}.$$
\end{ourlemma}
\begin{proof}
Let $r[\pfix{d}{i}]=\frac{1- \fhat{d}{i}}{3-p-(1-p)\cdot\fhat{d}{i-1}}$ and  $c[\pfix{d}{i-1}] = \frac{1-\fhat{d}{i-1}}{3-p-(1-p)\cdot\fhat{d}{i-1}}$ as it was defined in proof of Claim \ref{clm:ptam_dist}. We have,
\begin{align*}
    \frac{\Pr[\mal{D}'_i=d_i \mid \pfix{d}{i-1}]}{\Pr[D=d_i]} &= (1-p)+ p\cdot\left((c[\pfix{d}{i-1}])^k + \sum_{j\in [k-1]} (1-r[\pfix{d}{i}]) \cdot (1-c[\pfix{d}{i})]^j)\right)\\
    &=(1-p)+ p\cdot\left((c[\pfix{d}{i-1}])^k +  \frac{(1-r[\pfix{d}{i}])\cdot (1-c[\pfix{d}{i-1}]^k)}{1-c[\pfix{d}{i-1}]} \right).
\end{align*}
Also, in the proof of Claim \ref{clm:ptam_dist} we showed that,
$$\frac{\Pr[\mal{D}_i=d_i \mid \pfix{d}{i-1}]}{\Pr[D=d_i]} = 1-p + p\cdot\left(\frac{1-r[\pfix{d}{i}]}{1-c[\pfix{d}{i-1}]}\right).$$
Therefore, we conclude that,
\begin{align*}
    \frac{\Pr[\mal{D}'_i=d_i \mid \pfix{d}{i-1}]}{\Pr[\mal{D}_i=d_i \mid \pfix{d}{i-1}]} &= \frac{(1-p)+ p\cdot\left((c[\pfix{d}{i-1}])^k +  \frac{(1-r[\pfix{d}{i}])\cdot (1-c[\pfix{d}{i-1}]^k)}{1-c[\pfix{d}{i-1}]} \right)}{1-p + p\cdot\left(\frac{1-r[\pfix{d}{i}]}{1-c[\pfix{d}{i-1}]}\right)}\\
    &= 1 + \frac{p\cdot\left(\frac{(r[\pfix{d}{i}]-c[\pfix{d}{i-1}])\cdot c[\pfix{d}{i-1}]^k}{1-c[\pfix{d}{i-1}]} \right)}{1-p + p\cdot\left(\frac{1-r[\pfix{d}{i}]}{1-c[\pfix{d}{i-1}]}\right)}.
\end{align*}
We also know that $c[\pfix{d}{i-1}] \leq \frac{1}{2-p}$ because $\fhat{d}{i-1} \in [-1,+1]$. So we have,
\begin{align*}
    \frac{\Pr[\mal{D}'_i=d_i \mid \pfix{d}{i-1}]}{\Pr[\mal{D}_i=d_i \mid \pfix{d}{i-1}]}   &= 1 + \frac{p\cdot\left(\frac{(r[\pfix{d}{i}]-c[\pfix{d}{i-1}])\cdot c[\pfix{d}{i-1}]^k}{1-c[\pfix{d}{i-1}]} \right)}{1-p + p\cdot\left(\frac{1-r[\pfix{d}{i}]}{1-c[\pfix{d}{i-1}]}\right)}\\
    &\leq 1 + \frac{p\cdot c[\pfix{d}{i-1}]^k}{(1-p)\cdot(1-c[\pfix{d}{i-1}])} \\
    &\leq 1 + \frac{p}{(1-p)^2\cdot (2-p)^{k-1}} \leq \e^{\frac{p}{(1-p)^2(2-p)^{k-1}}}.
\end{align*}
Also for the inverse ratio, we have,
\begin{align*}
    \frac{\Pr[\mal{D}_i=d_i \mid \pfix{d}{i-1}]}{\Pr[\mal{D}'_i=d_i \mid \pfix{d}{i-1}]}   &= 1 + \frac{p\cdot\left(\frac{(c[\pfix{d}{i-1}]-r[\pfix{d}{i}])\cdot c[\pfix{d}{i-1}]^k}{1-c[\pfix{d}{i-1}]} \right)}{(1-p)+ p\cdot\left((c[\pfix{d}{i-1}])^k +  \frac{(1-r[\pfix{d}{i}])\cdot (1-c[\pfix{d}{i-1}]^k)}{1-c[\pfix{d}{i-1}]} \right)}\\
    &\leq 1 + \frac{p\cdot c[\pfix{d}{i-1}]^k}{(1-p)\cdot(1-c[\pfix{d}{i-1}])}\\
    &\leq 1 + \frac{p}{(1-p)^2\cdot (2-p)^{k-1}} \leq \e^{\frac{p}{(1-p)^2\cdot (2-p)^{k-1}}}.
\end{align*}
Therefore, we can finally conclude that,
$$\left|\ln\left(\frac{\Pr[\mal{D}_i=d_i \mid \pfix{d}{i-1}]}{\Pr[\mal{D}'_i=d_i|\pfix{d}{i-1}]}\right)\right| \leq \frac{p}{(1-p)^2 \cdot (2-p)^{k-1}}.$$
\end{proof}

\begin{ourlemma}\label{lemma:idealKcutBias}
Let $\mal{S}=(\mal{D}_1,\dots,\mal{D}_n)$ be the joint distribution after $\adv_\tam$ attack is performed on $S \equiv D^n$ using ideal $\pTam$ tampering algorithm. Also, let $\mal{S}'=(\mal{D}'_1,\dots,\mal{D}'_n)$  be the joint distribution after $\adv_\tam$ attack is performed on $S$ using Ideal $k$-cut $\pTam$ tampering algorithm where $k=\frac{\ln(2-p)-2\ln((1-p)\cdot \xi)}{\ln(2-p)}$. Then, it holds that,
$$\Ex[f(\mal{S'})]\geq \Ex[f(\mal{S})] -\e^{n\cdot\xi}+1.$$
\end{ourlemma}

\begin{proof}
Using Lemma \ref{lemma:kcut-dist}, for every prefix $\pfix{d}{i} \in \Supp(D)^i$ we have,
\begin{align*}
    \left|\ln\left(\frac{\Pr[\mal{D}_i=d_i \mid \pfix{d}{i-1}]}{\Pr[\mal{D}'_i=d_i|\pfix{d}{i-1}]}\right)\right| &\leq \frac{p}{(1-p)^2\cdot (2-p)^{k-1}} \leq \xi.
\end{align*}
Now, using Lemma \ref{lemma:closeAverage} we get
$\Ex[f(\mal{S'})]\geq \Ex[f(\mal{S})] -\e^{n\cdot\xi}+1.$
\end{proof}
We can now describe the actual efficient variant of our Ideal $\pTam$ attack.
\begin{ourconstruction}[Polynomial-time $k$-cut $\pTam$]\label{const:EffKcut}
Efficient $k$-cut $\pTam$ is the same as Ideal $k$-cut $\pTam$ of Construction \ref{const:IdealKcut} but it  it calls the semi-ideal oracle $\tilde{f_\xi}[\cdot]$ instead of the ideal oracle $\hat{f}[\cdot]$.
\end{ourconstruction}

\begin{ourlemma} \label{lem:approxR2} 
Let $0 < p < 1$. 
Let $0 < \xi < 1$.
Let $\alpha, \beta \in [-\xi, \xi]$.
Let $\fhat{d}{i-1}, \fhat{d}{i}, \afhat{d}{i-1}, \afhat{d}{i} \in [0, 1]$ such that 
$\afhat{d}{i-1} = \fhat{d}{i-1} + \alpha$
and
$\afhat{d}{i} = \fhat{d}{i} + \beta$.
Let $r[.]$ and $\tilde{r}[.]$ respectively be the rejection probabilities of the Ideal and Efficient $k$-cut $\pTam$. Then, for every $\pfix{d}{i} \in \Supp(D)^i$ we have
$|r[\pfix{d}{i}]-\tilde{r}[\pfix{d}{i}]| \leq O(\xi)$.
\end{ourlemma}
\begin{proof}
The proof is similar to the proof of Lemma \ref{lem:approxR}.
We have, 
\begin{align*}
    \left|r[\pfix{d}{i}]-\tilde{r}[\pfix{d}{i}]\right|
    &= \left|\frac{1- \fhat{d}{i})}{3-p-(1-p)\fhat{d}{i-1}} - \frac{1- \afhat{d}{i}}{3-p-(1-p)\afhat{d}{i-1}}\right| \,,
\end{align*}
where we can compute the following for the right hand side,  
\begin{align*}
    &= \left|\frac{(1-\fhat{d}{i})(3-p-(1-p)\afhat{d}{i-1}) - (1-\afhat{d}{i})(3-p-(1-p)\fhat{d}{i-1})}{(3-p-(1-p)\fhat{d}{i-1})(3-p-(1-p)\afhat{d}{i-1})}\right| \\
    &\le \left|\frac{(1-p)(\fhat{d}{i-1} - \afhat{d}{i-1}) + (3-p)(\afhat{d}{i} - \fhat{d}{i}) + (1-p)(\afhat{d}{i-1}\fhat{d}{i} - \fhat{d}{i-1}\afhat{d}{i})}{(3-p-(1-p))(3-p-(1-p))}\right| \\
    &\le \frac{(1-p)\xi + (3-p)\xi + (1-p)\left|\left(\fhat{d}{i-1}+\alpha\right)\fhat{d}{i} - \fhat{d}{i-1}\left(\fhat{d}{i}+\beta\right)\right|}{4} \\
    &\le \frac{4\xi + |\alpha| + |\beta|}{4} \le 3\xi/2\,.\qedhere
\end{align*}
\end{proof}

\paragraph{Putting things together.} 
Now we show how to choose the parameters of the Efficient $k$-cut $\pTam$. 
Suppose $\xi'$ is the parameter of Theorem \ref{thm:main-form}. 
If we choose $\xi$ as the parameter of our attack we can bound the final bias as follows. Firstly, if the approximation algorithm of Lemma \ref{lemma:approx_fhat} gives us a semi-ideal oracle $\tilde{f_\xi}[.]$, then based on Lemma \ref{lem:approxR2} we can approximate the rejection probabilities with error at most $O(\xi)$. 
Then based on  Lemma \ref{lemma:ImplicationOfApprox} the attack $\Adv_\tam$ that uses the efficient $k$-cut $\pTam$ generates a distribution that is $O(\frac{p}{1-p}\cdot k^2\cdot\xi)$-close to the distribution of the attack $\Adv_\tam$ that uses ideal $k$-cut $\pTam$. 

We use Lemma \ref{lemma:closeAverage} to argue that the bias of an efficient adversary is $\big(\e^{O(n\cdot \xi \cdot k^2 \cdot  \frac{p}{1-p})}-1\big)$-close to the bias of the ideal adversary. 
Also note that, if the approximation algorithm fails to provide a semi-ideal oracle for all queries, then bias of efficient attack is at least $-2$ because the function range is $[-1,+1]$. 
However, the probability of this event is bounded by $O(k\cdot n\cdot\xi)$ 
because the adversary needs at most $(k+1) \cdot n$ number of queries to $\tilde{f}$. 
Therefore, the difference of bias of the efficient and the ideal adversary is at most $O(k\cdot n\cdot \xi) + \e^{O(k^2\cdot n \cdot \xi\cdot \frac{p}{1-p})}-1$ which is at most $O(n\cdot \xi + k^2 \cdot n\cdot \xi\cdot \frac{p}{1-p})$ if the exponent in $\e^{O(k^2 \cdot n\cdot \xi\cdot \frac{p}{1-p})}$ is at most $1$. 
As a result, if we choose $\xi = o( \xi'/(k^2 \cdot n\cdot  \frac{p} {1-p})) = o(\xi' \cdot (1-p) / (k^2 \cdot n\cdot p))$, we can indeed guarantee that the bias of the efficient adversary 
(that uses efficient $k$-cut $\pTam$ tampering algorithm) is $\xi'$-close to the bias of the ideal adversary (that uses ideal $k$-cut $\pTam$). 

Now we want to select our other parameter $k$. 
Based on Lemma $\ref{lemma:idealKcutBias}$, if we choose $k=\omega \left(\frac{\ln((1-p)\xi')}{\ln(2-p)} \right)$ the bias of the attack $\Adv_\tam$ that uses the ideal $k$-cut $\pTam$ would be $\xi'$-close to the bias of the attack $\Adv_\tam$ that uses the ideal $\pTam$. 
Therefore, the bias of the $\Adv_\tam$ that uses efficient $k$-cut attack 
is $2\cdot\xi'$-close to the bias of $\Adv_\tam$ that uses ideal $\pTam$.

\section{Feasibility of PAC Learning under \ensuremath{p}-Tampering and \ensuremath{p}-Budget Attacks} \label{sec:PAC}
In this section, we study the non-targeted case where PAC learning could be defined.
We show that realizable problems that are PAC learnable (without attacks), 
are usually PAC learnable under $p$-tampering attacks as well. 
Essentially we 
bound the probability of some bad event happening (see Definition \ref{def:specialPAC}) 
in a manner 
similar to Occam algorithms \citep{OccamsRazor} by relying on the realizability assumption and relying on the specific property of the $p$-tampering attacks. In particular, we crucially rely on the fact that any $p$-tampering distribution $\mal{D}$ of a distribution $D$ contains a $(1-p)\cdot D$ measure in itself. In fact, we show (see Theorem \ref{thm:NoPAC-strong}) that in a close scenario to $p$-tampering in which the adversary can choose the ($\leq p$ fraction of the) tampering locations, PAC learning might suddenly become impossible. This shows that the `mistake-free' nature of $p$-tampering is indeed \emph{not} enough for PAC learnability.\footnote{We note that bounded-budget noise and in fact malicious has also been discussed outside of PAC learning; e.g.,~\citep{AMST:MaliciousErrors} in the membership query model of Angluin \citep{Angluin:ConceptLearning}.}

\subsection{Definitions}


\begin{ourdefinition}\label{def:bad}
For problem $\problem=(\XX,\YY,\distC,\hypoC,\loss)$, distribution $D \in \distC$, and training sequence  $\cS = ((x_1,y_1),\dots,(x_n,y_n)) \gets D^n$, 
we say that the event $\Bad_\eps(D,\cS)$ holds,
if there exists an $h \in \hypoC$ such that $h(x_i)=y_i$ for every $i \in [n]$ and $\Risk_D(h) > \eps$.
\end{ourdefinition}

\begin{ourdefinition}[Special PAC Learnability] \label{def:specialPAC}
A realizable problem $\problem=(\XX,\YY,\distC,\hypoC,\loss)$ is called \emph{special} $(\eps(n),\delta(n))$-PAC learnable  
if for all  $\dist\in\distC, n\in \N$, $\Pr_{\substack{\cS \gets D^n}}[\Bad_\eps(D,\cS)] \leq \delta(n).$
Special $(\eps(n),\delta(n))$-PAC learnability under poisoning attacks is defined similarly, where we demand the inequality to hold for every $\adv \in \advC_D$ tampering with the training set $\mal{\cS} \gets \adv(\cS)$. 
\end{ourdefinition}

It is easy to see that if $\problem$ is special $(\eps(n),\delta(n))$-PAC learnable, then it is $(\eps(n),\delta(n))$-PAC learnable through a `canonical' learner $L$ who simply finds and outputs a hypothesis $h$ consistent with the training sample set $\cS$. Such an $h$ always exists due to the realizability assumption. In fact, many \emph{efficient} PAC learning results follow this very recipe.\footnote{For example, properly learning monomials \citep{Valiant:PAC}, or using 3-CNF formulae to learn 3-term DNF formulae \citep{PittValiant::Hardness};  
the latter is an example of realizable but not proper learning.
As an example where the realizability assumption does not necessarily hold, see e.g., \citep{Diochnos::ALT2016},
for learning monotone monomials under a class of distributions - including uniform.
} 
That motivates our next definition. 

\begin{ourdefinition}[Efficient Realizability]
We say that the problem $\problem=(\XX,\YY,\distC,\hypoC,\loss)$ is \emph{efficiently} realizable, if there is a polynomial-time algorithm $M$, such that for all $\dist \in \distC$, and all  $\cS \gets D^n$, $M(\cS)$ outputs some $h \in\hypoC$ such that $\Risk_\dist(h) = 0$. 
\end{ourdefinition}

Here we define two types of tampering attackers who \emph{do} have control over which examples they tamper with, yet with a `bounded budget' limiting the number of such instances. Our definitions are inspired by the notions of \emph{adaptive corruption} \citep{STOC:CFGN96} and \emph{strong} adaptive corruption defined by Goldwasser, Kalai, and Park \citep{goldwasser2015adaptively} in the secure multi-party (coin-flipping) protocols.
\begin{ourdefinition}[$p$-budget attacks] \label{def:pBud} The class of \emph{strong} $p$-budget (tampering)  attacks $\advC^p_{\bud}$ $ = \cup_{D \in \cD} \advC_D$ is defined as follows. For  $D \in \distC$, any $\adv \in \advC_D$ has a (randomized) tampering algorithm $\Tamp$ such that:
\begin{enumerate}
\item Given access to a sampling oracle for distribution $D$, $\Tamp^D(\cdot)$ always outputs something in $\Supp(D)$.
\item Given any training sequence $\cS=(d_1,\dots,d_n)$,  the tampered output $\mal{\cS}=(\mal{d}_1,\dots,\mal{d}_n)$ is generated by $\adv$ inductively (over $i \in [n]$) as $ \mal{d}_i \gets \Tamp^D(1^n,\mal{d}_1,\dots,\mal{d}_{i-1},d_{i})$.
\item The number of locations that $\Tamp$ actually changes $d_i$ is bounded as $|\{i \mid d_i \neq \mal{d}_i \}| \leq p \cdot n$.
\end{enumerate}
\emph{Weak} $p$-budget tampering attacks are defined similarly, with the following difference. The tampering algorithm's 
execution $\Tamp^D(1^n,\mal{d}_1,\dots,\mal{d}_{i-1})$ 
is \emph{not} given $d_i$, but instead it could either output $o_i  \in \Supp(D)$, in which case we let $\mal{d}_i = o_i$, or it outputs a special symbol $\bot$, in which case we will have $\mal{d}_i = d_i$. Finally, since the weak $p$-budget attacker should make its decisions without the knowledge of $d_i$, we shall have $|\{i \mid \bot \neq {o}_i \}| \leq p \cdot n$.\footnote{The reason that we did not use the condition $|\{i \mid d_i \neq \mal{d}_i \}| \leq p \cdot n$ is the weak $p$-budget case is that, if the attacker chooses to tamper with the $i$'th location and simply happens to pick the same $o_i=d_i$, it should still count against its total budget.}
\end{ourdefinition}

\subsection{Our Results}

We first prove that PAC learning is possible under weak $p$-budget  (poisoning) attacks. We then show that this implies a similar possibility result under $p$-tampering attacks. We then prove that a similar result does \emph{not} hold for \emph{strong} $p$-budget  attacks in general. Our positive result (Theorem \ref{thm:PossPAC-weak}) holds even if the tampering algorithm is given all the history of tampered and untampered blocks (i.e., it is given given input $(1^n,\mal{d}_1,\dots,\mal{d}_{i-1},d_1,\dots,d_{i})$), and our impossibility result (Theorem \ref{thm:NoPAC-strong}) holds even if the tampering algorithm is given only $d_i$.

\begin{ourtheorem}[PAC learning under weak $p$-budget attacks] \label{thm:PossPAC-weak}
For any  $p \in (0,1)$, if a realizable problem  $\problem=(\XX,\YY,\distC,\hypoC,\loss)$ is $\left(\eps(n),\delta(n)\right)$-special PAC learnable, then, $\problem$ is  also $\left(\eps\left(n\cdot(1-p)\right),\delta\left(n\cdot(1-p)\right)\right)$-special PAC learnable under \emph{weak} $p$-budget (poisoning) attacks.
\end{ourtheorem}
\begin{proof} 
Without loss of generality, we can assume that the tampering algorithm of the adversary is deterministic (otherwise, we can fix the randomness to what is the best for the adversary, and we get a deterministic one again.) For $i\in [n]$ let $D_i$ be the random variable corresponding to the $i$th example before performing the tampering algorithm and let  $(\hat{D}_1,\dots ,\hat{D}_n)$ be the joint distribution of the training sequence after performing the tampering algorithm. 
Also let $T_i$ be a boolean random variable which is equal to 1 if the adversary picks to choose the $i$'th example and $T_i = 0$ otherwise. Using the notation of Definition \ref{def:pBud}, $T_i=0$ if $o_i=\bot$, and $T_i=1$ otherwise. 
For $i \in [(1-p)\cdot n]$ let $U_i$ be the random variable corresponding to the index of the $i$'th zero in the sequence $T_1,\dots,T_n$, and let $W_i \equiv \hat{D}_{U_i}$. 
We prove that the joint distribution $(W_1,\dots, W_{(1-p)\cdot n})$ is  distributed identically to $D^{(1-p)\cdot n}$. For every $i\in [(1-p)\cdot n]$ and $\pfix{d}{i}\in \Supp(D^i)$ we have, 
\begin{align*}
    \Pr[W_i = d_i \mid \pfix{W}{i-1} = \pfix{d}{i-1} ] = \sum_{j=1}^{n} \Pr[\hat{D}_j = d_i \mid \pfix{W}{i-1} = \pfix{d}{i-1} \wedge U_i = j]\cdot \Pr[U_i = j]
\end{align*}
\remove{
\Dnote{start the index of $j$ at 1 instead of 0?} \Mnote{I agree}\Snote{Fixed.}
 \Mnote{We say some sentences (before here) that are correct, but then we don't use them. better to say them when we use them.}\Snote{I moved the sentences.}
\Dnote{$\Pr[\hat{D}_i = D_i \mid T_i = 0]=1$ ? The rationale is that $\hat{D}_i$ is the output, so let's put it on the left.} \Mnote{I changed it. for things like this, just change them directly.}
}
Based on the assumption that the tampering algorithm of the adversary is deterministic, we know that $T_i$ is a function of $\pfix{D}{i-1}$. On the other hand, $D_i$ is independent of $\pfix{D}{i-1}$, so $D_i$ and $T_i$ are independent. Therefore, for all predicates $R\colon \Supp(\pfix{D}{i-1}) \to [0,1]$ such that $R(\pfix{D}{i-1})=1$ implies $T_i=0$ (i.e., $\Pr[T_i = 0 \mid  R(\pfix{D}{i-1}) = 1] = 1$) we have,
$$ \Pr[\hat{D}_i = d \mid R(\pfix{D}{i-1})=1] = \Pr[D_i = d \mid R(\pfix{D}{i-1})=1] = \Pr[D_i = d].$$
It is clear that $ \pfix{W}{i-1} = \pfix{d}{i-1} \wedge U_i = j$ is a predicate of $\pfix{D}{j-1}$ as it is a predicate of $\pfix{\hat{D}}{j-1}$ and $\pfix{T}{j}$. Also this predicate implies $T_j = 0$, therefore we have,
\begin{align*}
    \Pr[W_i = d_i \mid \pfix{W}{i-1} = \pfix{d}{i-1} ] &= \sum_{j=1}^{n} \Pr[\hat{D}_j = d_i \mid \pfix{W}{i-1} = \pfix{d}{i-1} \wedge U_i = j]\cdot \Pr[U_i = j]\\
    &= \sum_{j=1}^{n} \Pr[D_j = d_i]\cdot \Pr[U_i = j]= \Pr[D=d_i]
\end{align*}
which implies $(W_1,\dots,W_{(1-p)\cdot n}) \equiv D^{(1-p)\cdot n}$. 

Now let $\hat{\eps}(n)=\eps\left((1-p)\cdot n\right)$ and $\hat{\delta}(n)= \delta\left((1-p)\cdot n\right)$. Consider two sets,
$$\Good_1 = \{\cS \in \Supp(D^n) \colon \overline{\Bad_{\hat{\eps}(n)}(D,\cS)}  \} \text{ and } \Good_2 = \{\cS \in \Supp(D^{(1-p)\cdot n}) \colon \overline{\Bad_{\hat{\eps}(n)}(D,\cS)}\}.$$ 
Based on the definition of the event $\Bad$ (Definition \ref{def:bad}) we know that,
$$\Pr\left[(\hat{D}_1,\dots,\hat{D}_n) \in \Good_1 \mid (W_1,\dots,W_{(1-p)\cdot n})\in \Good_2\right]= 1.$$
Therefore we have,
\begin{align*}
\Pr\left[(\hat{D}_1,\dots,\hat{D}_n) \in \Good_1\right] &\geq \Pr\left[(W_1,\dots,W_{(1-p)\cdot n})\in \Good_2\right]\\
&= \Pr[D^{(1-p)\cdot n} \in \Good_2] \geq 1- \hat{\delta}(n).
\end{align*}
\end{proof}

We now derive the following theorem about $p$-tampering attacks from Theorem \ref{thm:PossPAC-weak}.

\begin{ourtheorem}[PAC learning under weak $p$-tampering attacks]  \label{thm:PossPAC-pTam}
For any  $p \in (0,1)$, if a realizable problem  $\problem=(\XX,\YY,\distC,\hypoC,\loss)$ is $(\eps(n),\delta(n))$-special PAC learnable, then for any $q \in (0, 1-p)$, $\problem$ is also  $(\eps'(m),\delta'(m))$-special PAC learnable under $p$-tampering poisoning attacks for $\eps'(m) = \eps(m\cdot(1-p-q)), \delta'(m) = \e^{-2m\cdot q^2}+\delta(m\cdot(1-p-q))$. Thus, if $\problem$ is efficiently realizable and  special PAC learnable, then $\problem$ is also  efficiently PAC learnable under $p$-tampering.
\end{ourtheorem}

\begin{proof}
Consider a $p$ tampering attacker. By Hoeffding inequality of Lemma \ref{lem:hoeffding}, the probability that this attacker tampers with more than $(p+q)\cdot m$ input instances is at most $\e^{-2m\cdot q^2}$. Therefore, with probability $1-\e^{-2m\cdot q^2}$, this attacker is a \emph{special} case of a weak $(p+q)$-budget attacker, as it does \emph{not} choose the locations of the attack, and thus cannot choose the tampering locations based on the content of the training examples. Therefore, we can obtain the same bounds of Theorem \ref{thm:PossPAC-weak}, but we shall use $p+q$ as the budget (fraction) and also add $\e^{-2m\cdot q^2}$ to the confidence error.
\end{proof}

\begin{ourtheorem}[Impossibility of PAC learning under strong $p$-budget attacks] \label{thm:NoPAC-strong}
For any constant $p \in (0,1)$, there is a problem
$\problem=(\XX,\YY,\distC,\hypoC,\loss)$ that is PAC learnable (under no attack), but it is not PAC learnable under strong $p$-budget  (poisoning) attacks.
\end{ourtheorem}
\begin{proof}
Suppose $\XX=[k]$ where $k = \ceil{\frac{2}{p}} $. Let  $\YY = \set{0,1}$, and suppose $\cD$ consists of all $(x,c(x))_{x \gets \XX}$  where $x \gets \XX$ is an example drawn from $\XX$ uniformly at random and $c$ is an arbitrary  function (concept) in $\YY^\XX$. Let the hypothesis class $\hypoC$ contain all of $\YY^\XX$, 
and $\Loss(b_0,b_1) = |b_0-b_1|$ is the natural loss for classifiers. 
%

PAC learnability of $\problem$ trivially follows from the fact that $|\XX|=k$ is finite. Therefore, enough samples will reveal the concept function $c$ (defined through $D$) completely with overwhelming probability for large enough samples $n$. Consider a concept class which consists of only two functions $c_0$ and $c_1$ such that,
$$c_0(i) = 0, \forall i\in [k], \text{ and }$$
\begin{align*}
c_1(i) = 
\begin{cases} 
      0 & i \in [k-1] \\
      1 & i = k.
   \end{cases}
\end{align*}
Now we propose a strong $p$-budget adversary $A_{sb}$ ($sb$ stands for \textit(strong budgeted))  that replaces every pair $(k,*)$ it sees with $(k-1,0)$ until it runs out of its budget which is $p\cdot n$ examples.  
\remove{
\Dnote{Perhaps replace $(k, c_i(k))$ (for $i\in\{0, 1\}$) with $(k-1, 0)$? This way the example $(k-1, 0)$ is `correct' regardless if the target function is $c_0$ or $c_1$.
Moreover, the adversarial distribution that vanishes $k$ (whp), is the same as we want to prove afterwards.
In other words, if the instance is $k$, we ignore the label and we substitute with another element of the domain.
(By the way, this method has flavor very close to the idea of the induced distributions - though, we never really set some adversarial distribution, but work with what we have ...)} \Mnote{I agree, we should substitute $(k,*)$ with $(k-1,0)$. Is it identical to what they do? We are also choosing the distribution to be uniform. Can we use their result as a blackbox?} \Dnote{No. What we do is still different. The idea is similar in the sense that we can not `disambiguate' among two different concepts. Other than that, our approach is different. Their distribution is highly non-uniform. :)} \Snote{Good point! I changed it.}
}
We denote the distribution of examples after the attack is performed by $A_{sb}(D^n)$. Let us define an event $\sfE$ which is $0$ if the adversary runs out of budget at some point and is $1$ if she does not run out of budget. Note that if $c_0$ is being used then the adversary will not do any thing at all and cannot run out of budget. 
If $c_1$ is used we can bound the probability of adversary running out of its budget using Chernoff bound  as follows,
$$\Pr[\sfE] \geq 1 - \e^{\frac{-n}{3k}}.$$
\remove{
\Dnote{I think we mean the adversary will \emph{not} run out of budget with at least that much probability.} \Mnote{But that is what E is defined to be, no?} \Dnote{You are right. I misread the sentence. I am sorry.}}
Let $L$ be a learning algorithm that is going to learn a concept $c$ sampled uniformly from $\{c_0,c_1\}$ by looking at $n$ labeled examples sampled from $A_{sb}(D_c^n)$ where $D_c \equiv (d, c(d))_{d \gets U_{[k]}}$ . We have,
$$\Pr_{\substack{c\gets \{c_0,c_1\}\\h \gets L(A_{sb}(D_c^n))}}[h(k) = c(k) \mid\sfE] \leq \frac{1}{2}.$$
The reason is that two conditional distributions $(A_{sb}(D^n_{c_0}) \mid \sfE)$ and $(A_{sb}(D^n_{c_1}) \mid \sfE)$ are identical, and there is no way for the learning algorithm to find out which of these distributions are being used. Therefore,
\begin{align*}
\Ex_{\substack{c\gets \{c_0,c_1\}\\h \gets L(A_{sb}(D_c^n))}}[\Risk_{D_c}(h)] &\geq
\frac{1}{k}\cdot\Pr_{\substack{c\gets \{c_0,c_1\}\\h \gets L(A_{sb}(D_c^n))}}[h(k)\neq c(k)] \\
& \geq \frac{1}{k}\cdot \Pr_{\substack{c\gets \{c_0,c_1\}\\h \gets L(A_{sb}(D_c^n))}}[h(k)\neq c(k) \mid \sfE]\cdot \Pr[\sfE]\\
&\geq \frac{1-\e^{\frac{n}{3k}}}{2k}.
\end{align*}
Now let $\eps_c(n)$ and $\delta_c(n)$ be the error and confidence that $L$ provides when using $n$ examples sampled from $A(D_c^n)$. We know that,
$$\Ex_{h \gets L(A_{sb}(D_c^n))}[\Risk_{D_c}(h)] \leq \eps_c(n) + \delta_c(n)$$
which implies,
$$\Ex_{\substack{c\gets \{c_0,c_1\}\\h \gets L(A_{sb}(D_c^n))}}[\Risk_{D_c}(h)] \leq \frac{\eps_{c_0}(n) + \delta_{c_0}(n) + \eps_{c_1}(n) + \delta_{c_1}(n)}{2}.$$
Therefore we have, 
$$\eps_{c_0}(n) + \delta_{c_0}(n) + \eps_{c_1}(n) + \delta_{c_1}(n) \geq \frac{1-\e^{\frac{-n}{3k}}}{k}$$
which means for any learning algorithm $L$, one of these values will remain at least $\Omega(1/k) = \Omega(p)$ no matter how many examples the algorithm uses.

\end{proof}
\remove{
\begin{proof}
Suppose we sample $\cS \gets D^m$. 
By a Chernoff bound, an adversary that tampers with each of the examples in $\cS$ independently with probability $p$, 
will not change more than a $p+q$ fraction of the elements of $\cS$ except with probability at most 
$\e^{-2mq^2}$.
Thus, with high probability, 
at least $(1-p-q) \cdot m \geq n$ 
examples 
in the tampered training sequence $\mal{\cS}$ are sampled from $\DD$ \emph{without} any control from the adversary.  \Mnote{This needs a more formal argument. It can be as follows. We first sample enough samples outside of the attack's setting, and whenever adversary's coin says there is no tampering happening, we use one of the unused examples that are chosen before. Maybe, we should skip a proof here and only get this as a black-box from the result of Theorem 5 below? It seems we can do use Theorem 5 + a Chernoff bound to say that we won't pick too many tampering locations.}
Since \problem is 
special $(\eps(n),\delta(n))$-PAC learnable, with probability at least $1-\delta(n)$, 
these $n$ `untampered' examples from $\DD$ will eliminate any hypothesis 
with 
risk larger than $\eps$.
Since the tampered sequence $\mal{\cS}$ of a $p$-tampering attack is 
in the support set $\Supp(D)^n$, 
due to realizability, 
there is at least one $h$ such that $\Risk_D(h)=0$.
Hence, the learner 
can still find and output at least one 
$h\in\hypoC$ 
for which $\Risk_D(h) \leq \eps$. 
If further, $\problem$ is efficiently realizable, 
$h$
can be found in polynomial time as well.
\end{proof}
}
\section{Open Questions}
We conclude with discussing some natural directions for future work that remain open following our work.

\paragraph{Bounds for attacking specific problems and/or specific learners.} 
The bounds of Corollaries \ref{cor:NonTar} and \ref{cor:Targ} 
apply to \emph{any} PAC learning problem $\problem$ and \emph{any} learner $L$ for problem $\problem$. Therefore, one can possibly get much stronger bounds for \emph{specific} learning problems, and even for a fixed learning problem $\problem$, one can get even better bounds if specific learning algorithms are attacked.

\paragraph{Learning under $p$-tampering without realizability.} The result of Theorems  \ref{thm:PossPAC-weak} and \ref{thm:PossPAC-pTam} require the realizability assumption to hold for the learning problem  $\problem$. In what settings do these result extend without the realizability assumption?

\paragraph{Learning under \emph{targeted} $p$-tampering.} Theorems \ref{thm:PossPAC-weak} and \ref{thm:PossPAC-pTam} both apply to the case of \emph{non-targeted} poisoning attacks, where the adversary does \emph{not} know the final test example. A natural open question is whether, at least for specific natural cases, this result extends even to the targeted case, where the adversary's tampering strategy could depend on the final test example drawn from the same distribution $D$ as that of training.

\paragraph{Complementary positive result for Theorem \ref{thm:NoPAC-strong}.} Based on Definition \ref{def:pBud}, the attacks of \citep{KearnsLi::Malicious} in the malicious noise model  fall into our category of weak attacks (as they do not need to know the label of the tampered example) while attacks in the nasty noise model  are of strong form. However, both of these works \citep{KearnsLi::Malicious,NastyNoise} are allowed to generate examples with wrong labels. Both of these works \citep{KearnsLi::Malicious} and \citep{NastyNoise} also prove lower bounds and matching upper bounds for the achievable accuracy of learners in presence of malicious noise and nasty noise respectively. Our Theorem \ref{thm:NoPAC-strong} proves a lower bound of $\omega(p)$ on the achievable accuracy or the confidence parameter of learners in presence of `strong' $p$-budgets attacks which are limited to use examples with correct labels. Are there any similar matching upper bounds for the lower bounds of Theorem \ref{thm:NoPAC-strong}?

\paragraph{}{\large  Acknowledgement.} We would like to thank the anonymous reviewers of ALT 2018 as well as ISAIM~2018 for their useful comments on earlier versions of this work.

\bibliographystyle{alpha}
\newcommand{\etalchar}[1]{$^{#1}$}

\appendix



\end{document}